\documentclass[10pt,twocolumn,letterpaper]{article}

\usepackage{cvpr}              % To produce the CAMERA-READY version
\usepackage{graphicx}
\usepackage{amsmath}
\usepackage{amssymb}
\usepackage{booktabs}
\usepackage{multirow}
\usepackage{soul}
\usepackage{xcolor}

\usepackage[pagebackref,breaklinks,colorlinks]{hyperref}

\usepackage[capitalize]{cleveref}
\crefname{section}{Sec.}{Secs.}
\Crefname{section}{Section}{Sections}
\Crefname{table}{Table}{Tables}
\crefname{table}{Tab.}{Tabs.}

\newcommand{\MUSE}{\texttt{MUSE-VAE}\xspace}

\begin{document}

\renewcommand*{\figureautorefname}{Fig.}
\renewcommand*{\tableautorefname}{Tab.}
\renewcommand*{\sectionautorefname}{Sec.}
\renewcommand*{\subsectionautorefname}{Sec.}
\renewcommand*{\equationautorefname}{Eq.}

\title{
MUSE-VAE: Multi-Scale VAE for Environment-Aware\\ Long Term Trajectory Prediction
}

\author{Mihee Lee\\
Rutgers University\\
USA\\
{\tt\small ml1323@rutgers.edu}
\and
Samuel S. Sohn\\
Rutgers University\\
USA\\
{\tt\small samuel.sohn@rutgers.edu}
\and
Seonghyeon Moon\\
Rutgers University\\
USA\\
{\tt\small sm2062@cs.rutgers.edu }
\and
Sejong Yoon\\
The College of New Jersey\\
USA\\
{\tt\small yoons@tcnj.edu}
\and
Mubbasir Kapadia\\
Rutgers University\\
USA\\
{\tt\small mubbasir.kapadia@rutgers.edu}
\and
Vladimir Pavlovic\\
Rutgers University\\
USA\\
{\tt\small vladimir@cs.rutgers.edu }
}
\maketitle

\begin{abstract}

Accurate long-term  trajectory prediction in complex scenes, where multiple agents (e.g., pedestrians or vehicles) interact with each other and the environment while attempting to accomplish diverse and often unknown goals, is a challenging stochastic forecasting problem.  In this work, we propose \MUSE, a new probabilistic modeling framework based on a cascade of Conditional VAEs, which tackles the long-term, uncertain trajectory prediction task using a coarse-to-fine multi-factor forecasting architecture.  In its Macro stage, the model learns a joint pixel-space representation of two key factors, the underlying environment and the agent movements, to predict the long and short term motion goals. Conditioned on them, the Micro stage learns a fine-grained spatio-temporal representation for the prediction of individual agent trajectories.  The VAE backbones across the two stages make it possible to naturally account for the joint uncertainty at both levels of granularity.    As a result, \MUSE offers diverse and simultaneously more accurate predictions compared to the current state-of-the-art. We demonstrate these assertions through a comprehensive set of experiments on nuScenes and SDD benchmarks as well as PFSD, a new synthetic dataset, which challenges the forecasting ability of models on complex agent-environment interaction scenarios.

% Predicting human trajectories and planning a safe path has become a crucial problem with the rise of the autonomous driving business. Learning a model that can predict realistic trajectories without collision with obstacles is essential. We propose MacMicHTP that approaches from Macro to Micro steps based on Conditional VAE (CVAE). In the Macro-stage, we process the trajectories in the pixel coordinate to achieve  environment-aligned learning without distortion of spatial signal. At this stage, our model learns the uncertainty of the long-term goal position based on CVAE framework. After learning the multimodal distribution of the goal position, in the Micro-stage, our model leverage RNN-based CVAE to predict the full sequential trajectories based on the predicted goal positions. We demonstrate our model on three datasets. First, we provide a simulated path-finding dataset to verify the effectiveness of our environment-aware model. We also perform experiments on two challenging real-world data sets, nuScenes and SDD, and demonstrate the performance significance.
\end{abstract}

\vspace{-0.2in}
\section{Introduction}
\label{sec:1}
Human behavior forecasting is an essential problem studied in various research fields such as computer vision~\cite{socialgan}, computer  graphics~\cite{helbing1995social}, robotics~\cite{ferrer2013social}, and  cognitive science~\cite{wiener2009taxonomy}. The fundamental problem with predicting human motion is the inherent stochasticity stemming from the fact that human beings use numerous sources of information to make a wide variety of different decisions at any given moment, which all impact their future movement. This movement uncertainty translates beyond the motion of the humans alone to the movement of objects controlled by humans, such as vehicles \cite{nusc}. 
% \commk{add citation}

To embrace the uncertainty, in this paper, we focus on developing computational models, learned from data, that can predict a realistic multi-modal distribution of the future agent (humans, vehicles, etc.) trajectories.  The models are designed in the context of two main  factors that drive this uncertainty: the environment the agents occupy and the task they are attempting to accomplish. %\commk{instead of future tense, switch to present perfect or similar. e.g. change ``will be'' to ``are'' }

\begin{figure}
\centering
% \begin{subfigure}[tbhp]{0.157\textwidth}
%  {\includegraphics[width=\textwidth]{figures/intro_lg3.png} 
%  \caption{Long-term goal} \label{fig:intro_lg} 
%  }
% \end{subfigure}
% \hfill
% \begin{subfigure}[tbhp]{0.312\textwidth}
%  {\includegraphics[width=\textwidth]{figures/intro_sg3.png} 
%  \caption{Intermediate Short-term goals} \label{fig:intro_sg} }
% \end{subfigure}

\vspace{-0.1in}
\begin{subfigure}[tbhp]{0.35\textwidth}
 {\includegraphics[width=\textwidth]{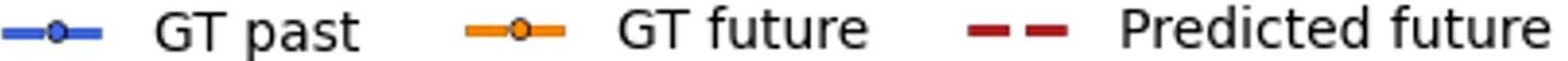} 
 }
\vspace{-1.4em}
\end{subfigure}
\begin{subfigure}[tbhp]{0.23\textwidth}
 {\includegraphics[width=\textwidth]{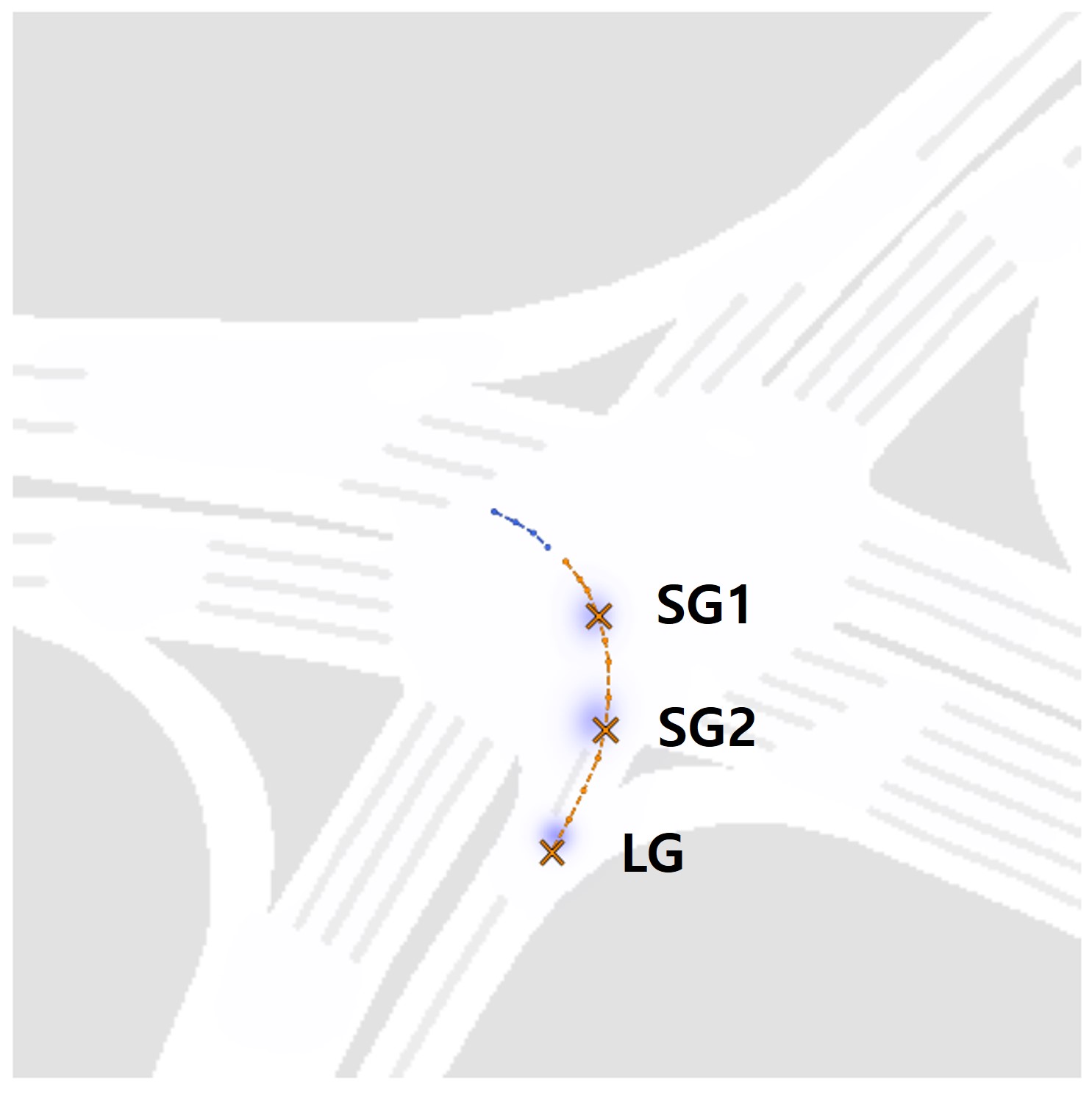} 
 \caption{Intermediate Short-term goals} \label{fig:intro_lsg} }
\end{subfigure}
\begin{subfigure}[tbhp]{0.24\textwidth}
 {\includegraphics[width=\textwidth]{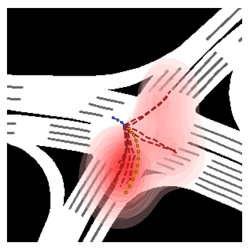} 
 \caption{Full trajectories} \label{fig:intro_micro} 
 }
\end{subfigure}
\vspace{-0.1in}
  \caption{ 
  (a) The predicted trajectory heatmaps are overlaid in the semantic map. Ground Truth (GT) long-term goal (LG) and short-term goals (SG1 and SG2) are marked with `x'.
  (b) Complete trajectory forecasting based on the predicted LG and SG. Each sequence of trajectories is obtained from a different pair of LG and SG predictions.}
\label{fig:intro}
\vspace{-1.5em}
\end{figure}

However, direct forecasting of long-term trajectories is a challenging task.  A person typically plans one’s movement in a coarse-to-fine fashion: with a final destination in mind, through a sequence of intermediate goals or way-points, the movement is executed to reach those sub-goals~\cite{chown1995prototypes,sharif2021subcircuits}.
% \comvp{would be great if we could support this by references}
% By narrowing the entire journey to the destination into a partial sequence of steps, we can set the target position of the sequence, the last step of the partial sequence. 
State-of-the-Art (SOTA) methods \cite{tnt, ynet, stepwisegoal} leverage this intuition to propose goal-conditioned prediction model. However, despite their effectiveness compared to traditional approaches~\cite{socialattn, socialgan, sociallstm}, these models show limited ability to deal with complex environments~\cite{stepwisegoal}, particularly as they affect the movement~\cite{tnt}. This often results in physically implausible trajectory predictions that violate agent-environment collision constraints. Moreover, the models frequently struggle to account for the diversity of the forecast goals and trajectories~\cite{ynet}, which are driven by the uncertain, multi-modal nature of the problem. %prediction problem. 
% After the goal position is set, a person will search for his way to the goal in the most efficient way avoiding the obstacles. 

%Inspired by the multi-scale nature of the human behavior, 
To address this, we propose \MUSE: a multi-scale, environment-aware model for long-term trajectory prediction which (1) takes a stage-wise, coarse-to-fine approach to trajectory prediction by predicting both the higher-level goals and the goal-conditioned trajectory, (2) avoids collision with obstacles without loss of spatial signal which can occur due to spatial reorganization when compressing 2D information into 1D features,
% \comvp{it is not clear what 'spatial signal' here means, the term was not defined or mentioned previously}
and (3) learns a multimodal predictive distribution across the stages, thus capturing the inherent uncertainty.
\MUSE embodies a three-step learning strategy across a Macro-stage and Micro-stage.
The Macro-stage comprises of two steps for coarse predictions. We first predict the long-term goal, i.e., the last step of the given sequence based on heatmap trajectory representation. Given the long-term goal, two sequential short-term goals are predicted as shown in \autoref{fig:intro_lsg}.
% \comvp{again, the terms image and world coordinates are given out of context here, undefined, and could mean anything. either introduce the terms when discussing the failure of the prior work or explain them here. I would suggest the former because that gives you the opportunity to motivate why pixel-level is needed, e.g., to easily model uncertainty in coarse goals on a discrete (image) grid.} 
After getting the goal positions in the Macro-stages, finally, our model produces the full trajectories in the Micro-stage as in \autoref{fig:intro_micro}. Our main contributions are as follows: 
(\textbf{a}) We introduce a novel multi-scale learning strategy for CVAE-based probabilistic models in order to make environment-aware collision-free trajectory predictions. 
(\textbf{b}) Unlike the prior works, we show that one can learn trajectory distributions that can be well generalized in new scenes at test time, giving various reasonable predictions compliant to the environment without needing extra steps for diversity. 
(\textbf{c}) The proposed coarse-to-fine approach enables diverse and accurate trajectory predictions by forecasting the heading of the entire trajectories through goal prediction and then expanding it to granular and complete predictions.

We demonstrate these contributions through experiments on both real and synthetic dataset. With various grounded evaluation metrics, we show that \MUSE can produce predictions similar to GT trajectories while achieving less collision with the environment than the SOTA methods.  

\vspace{-0.1in}
\section{Related Work}
\label{sec:related 2}
\vspace{-0.1in}
The modeling of agent movement behavior, including individual humans, crowds, vehicles, etc., is a long-standing problem crossing the boundaries of multi-agent and computer vision communities.  We focus on three relevant aspects: the forecasting of individual trajectories, the interplay between movement behavior and the environment, and the need for modeling thet uncertainty in motion prediction.
% \comvp{I would reorganize the subsections to align with the above ordering of topics.}

\noindent\textbf{Sequence Learning } The human trajectory has a sequence characteristic that changes in turn according to the passage of time. In order to capture the nature of the sequential information, many prior works \cite{sociallstm, socialgan, socialattn, sophie, t++, desire} utilize Recurrent Neural Networks (RNNs) \cite{rnn} such as LSTMs and GRUs. 
However, RNN suffers from forgetting the past hidden states as the recursion goes. 
\cite{Giuliari2021TransformerNF, AgentFormer} tackle the temporal aspects of human trajectory forecasting by adopting Transformer Networks \cite{transformer}. Transformer solves the long-range dependency problem by processing the a sequence as a whole with self-attention and  positional encoding. 
Y-net \cite{ynet} solves the sequential trajectory learning problem with only convolution layers. They represent trajectories with multiple heatmaps, which are stacked with the semantic environment map image along the channel dimension and fed to their convolution networks as a whole. This way, they learn temporal movements with the environment without tradition sequence learning networks.

\noindent\textbf{Environment Learning } A decision about the trajectory taken towards a goal depends on the surrounding environment. Many prior approaches provide environmental information to their model for realistic trajectory predictions. \cite{t++, AgentFormer, tnt, sophie} encode the environment layout and semantics as a representation of the scene image with a convolution network and use it to train their models along with trajectory features. While these approaches can learn the scene context surrounding the trajectory, they compress it into 1D feature vectors after CNNs and FCs layers, 
% \comvp{not clear what this 1d fv is or where it comes from} 
which can convey corrupted information in terms of spatial signals.
% \comss{Do we need to substantiate this?}
Y-net \cite{ynet} addresses this issue by aligning the semantic map with the trajectory heatmap spatially and processing them as a whole. Our model attempts more meaningful environmental learning without unnecessary information by focusing on a limited area around the trajectory rather than the entire scene while keeping the spatial signal by utilizing the heatmap trajectory representation.

\noindent\textbf{Multimodal Learning } The trajectory of an agent (human, vehicle, etc.) is affected by a number of factors such as the destination in mind, the surrounding environment, nearby agents and so on, which leads to an intrinsic uncertainty about the future behavior. Recent studies focus on learning the \textit{distribution} of the human trajectory based on deep generative models, sidestepping the deterministic trajectory prediction. \cite{t++, desire, AgentFormer, Tang2019MultipleFP, trajectron} adopt Conditional Variational AutoEncoders (CVAE) \cite{cvae} and \cite{socialgan, sBiGATMT, sophie} introduce Generative Adversarial Network (GAN) \cite{gan} for learning of trajectory distribution where multiple predictions can be sampled. 
Trajectron++ \cite{t++} tackles the multimodal aspect of trajectory distributions by adopting a discrete latent distribution for the latent space, and Gaussian Mixture Model as the output distribution of the decoder in their CVAE framework.
% \comvp{would be good to say what each one is for}. 
AgentFormer \cite{AgentFormer} promotes diversity of the predictions with a pairwise distance loss across predictions. However, this approach requires retraining whenever a different number of predictions are sought at test time. Y-net \cite{ynet} utilizes K-means clustering of predictive discrete density maps at test time to achieve diverse prediction; however, the model does not explicitly learn the resolution-free multimodal trajectory density.
Some prior works \cite{tnt, ynet, stepwisegoal, PRECOGPC} encourage the multimodality by proposing a goal-conditioned forecasting model under the assumption that one's movement depends primarily on the final goal position.
% We introduce a model that better reflects the characteristics of a past sequential properties such as velocity as well as achieves environment collision-free predictions.
% \comvp{realistic prediction is vague, should state what you mean by it; e.g., more accurate in final goal, more accurate in the whole trajectory, ...?}.

\MUSE adopts a stage-wise training procedure to incorporate sequential information 
% \comvp{term robust is vague here; what do you mean by it?} 
while maintaining a trajectory aligned with the environment.
First, in the Macro-stage, future predictions are obtained by utilizing the heatmap representation of trajectories along with the semantic environment map, and then in the Micro-stage, RNN-based networks are used to facilitate sequence learning.
The Micro-stage takes advantage of coarse predictions from the Macro-stage, reducing the long-range dependency problem and guiding the path to avoid obstacles. 
Adopting VAE in both Macro- and Micro-stages, our model learns the inherent uncertainty of forecasting, which can give a variety of plausible predictions.

\vspace{-0.1in}
\section{Proposed Method}
\label{sec:3}

\begin{figure}
\centering
\begin{subfigure}[tbhp]{0.47\textwidth}
 {\includegraphics[width=\textwidth]{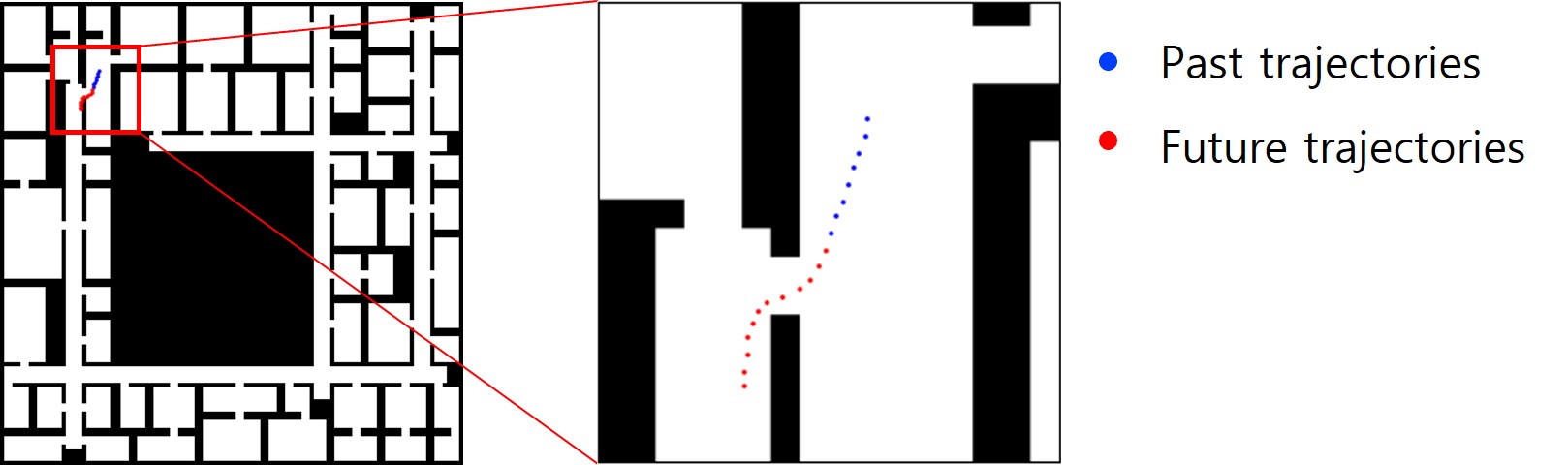} 
 \caption{Global / Local view of the semantic map} \label{fig:local_map} 
 }
\end{subfigure}

\begin{subfigure}[tbhp]{0.47\textwidth}
 {\includegraphics[width=\textwidth]{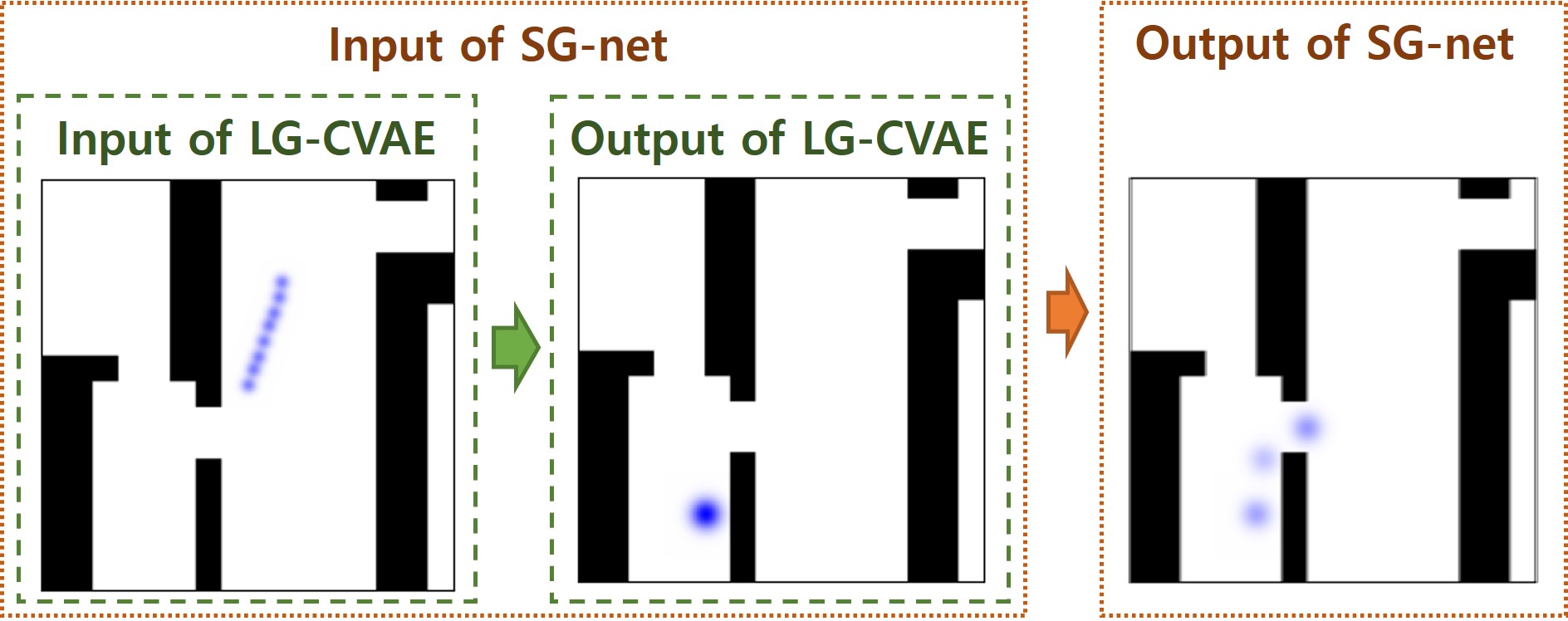} 
 \caption{Input / Output of the Macro-stage models} \label{fig:macro_inout} }
\end{subfigure}
  \caption{(a) The semantic map with 8 past / 12 future trajectories. Rather than the global map, we use the local map to focus on the nearby environment of the given trajectories. (b) Input and output format of Macro-stage models, LG-CVAE and SG-net. Trajectory heatmaps are overlaid with the local view semantic map. Here we assume 2 short-term goals at future time step 4 and 8 among 12 future steps. Thus, SG-net outputs 3 heatmaps; 2 for the short-term goals and 1 for the long-term goal.
%   \commk{mihee, it looks awkward to put (a) and (b) together since the aspect ratio of (a) gets shrunk. i suggest making these separate images } \comsy{next to (a), you can add legends for blue(past)/red(future) to fill up the space and align the figures}
% \comvp{Looks better, but I still find the heatmaps of LT and ST goals not prominent enough; I would manually edit the images to enlarge the spread/intensity in those maps, or just run another amplifying gaussian filter over those heatmaps.}
}
\label{fig:inout}
\vspace{-1.5em}
\end{figure}

\begin{figure*}[t]
\centering
{\includegraphics[width=0.95\textwidth]{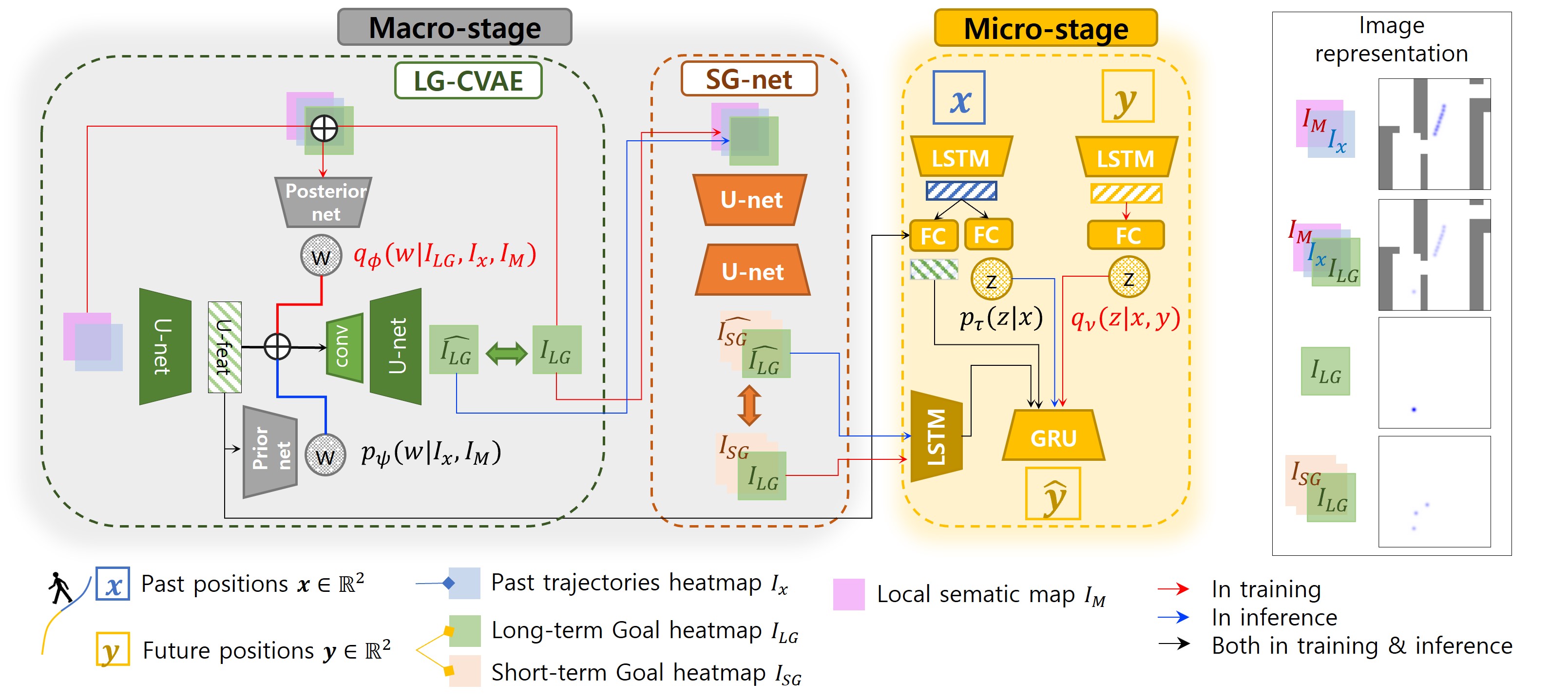}}
\vspace{-.1in}
\caption{\MUSE architecture. LG-CVAE is the first stage which predicts the long-term goal based on CVAE framework. Conditioned on the long-term goal, SG-net predicts the waypoints from the past trajectories to the long-term goal. We group these two stages as Macro-stage where the predictions are made in heatmap representation to keep the spatial signal along with the semantic map. Finally in Micro-stage, full trajectories are obtained with RNN-based CVAE.
More implementation details are in the Supplementary Materials.
% \comvp{Can we add depiction of x,y,hat y?  Eg, you can use smooth curves to illustrate that these entities are in real-valued world coordinates.}
% \comsy{suppressed comments to guage the page length}
%-------------------------------------------------------------
% unresolved comments
%-------------------------------------------------------------
%\comvp{I think it would also be useful/essential to depict that all input maps/images are constructed from their respective continuous valued entities, ie image $I_x$ is a deterministic mapping from $x$, image $I_m$ is another deterministic mapping from $S$, etc.  As is, one can get the impression that images and trajectories are unrelated/independent.  That is also not formalized in the Methods section itself, but it should be.}
%\comml{the mapping for $I_M$ is described in right above LG-CVAE subsection because it needs more notations to be explained}
%-------------------------------------------------------------
}
\label{fig:arch}
\vspace{-1.em}
\end{figure*}

The trajectory prediction problem is formulated as follows. Assume that we are given $t_p >0$ timestamps, the past trajectory positions 
$ x =  \{x_i^t\}_{t={1}}^{t_p} $ 
% $ \mathcal{X} =  \{x_i^t\}_{t={1}}^{t_p} $ 
% \comvp{what is $t_1$, it is not defined?} 
of agent $i$ in scene $S$, where $x_{i}^{t} \in \mathbb{R}^{2}$ denotes the 2D world coordinates of the agent $i$ at time $t$. Our goal is to predict the future trajectory of the same agent during $t_f >0$ future timestamps,  $ y = \{y_i^t\}_{t=t_{p+1}}^{t_{p+f}}$ in the sense of their distribution. $y_{i}^{t} \in \mathbb{R}^{2}$ is the future 2D position in the same coordinate system as $x_{i}^{t}$.
% \comvp{should formally define what each trajectory step means, ie location or loc+velocity etc. This is also the place where you need to define most terms/variables, eg S is specified here but never used again.  You use M later.  Maybe that's ok, but you need to then later say that M is a semantic map of env S and how it is represented.  Also, is there a particular reason to use x and y with different time indexes to represent the same quantity, i.e., agent location?  In other words, why is future also not denoted by x?}. 
% \comml{M is defined in sec 3.1. I used y rather than x to indicate the future trajectories to make representation easier in section 3 - figures and equations }
% \comvp{I still see a lot of inconsistencies in the notation used. This may be the source of confusion when one reads the paper.  Please make an effort to use consistent notation, which will be used and re-used in the rest of the paper.  And still it is not clear what $x_i^t$ is: position, pos+velocity, ...?}
%\comvp{I think I added this comment three times so far: we NEED TO PRECISELY DEFINE what $x_i^t$ is: 2D position in real-valued world coordinates, pos+velocity, something else,...}
%\comml{I wrote it as trajectory positions $ x \{x_i^t\}_{t={1}}^{t_p} $  above. should I mention it as 2D position in real world $ x =  \{x_i^t\}_{t={1}}^{t_p} $ ? }
%\comsy{hopefully this clarifies..}
This prediction should take into account the environmental context $S$, i.e., $p(y \vert x, S)$. 
%\comvp{just to verify: the model does not take into account other agents?}
%\comml{Our model does not consider agent interaction}
We propose our MUlti-Scale Environment-aware model, \MUSE for coarse-to-fine trajectory forecasting.
The Macro-stage is defined as a coarse prediction of the future trajectories, and the Micro-stage is defined as a fine prediction based on the coarse prediction. 
In the Macro-stage, only a subset of the future steps are predicted as the long-term and short-term goals. We denote the long-term goal as the final step at $t_{LG} = t_{p+f}$ and the short-term goals as some intermediate steps $ t_{SG} \in \{ t_{p+1}, \ldots, t_{p+f-1}\}$. The Macro-stage aims to obtain rough predictions that are well aligned with the scene for collision avoidance against environmental obstacles.
Based on the coarse prediction, the Micro-stage generates a fine-grained prediction of all $t_f$ future steps. In this stage, we adopt the RNN \cite{rnn} to efficiently learn the sequential features of trajectories.

In \autoref{sec3.1.macro}, we introduce the coarse prediction stage, Macro-stage,  and elaborate on how the primary Macro-stage model, Long-term Goal Conditional VAE (LG-CVAE), and the subsequent Macro-stage model, Short-term Goal network (SG-net), are formulated. \autoref{sec3.2.micro} introduces the Micro-stage, the fine prediction stage, used to refine predictions of complete forecast trajectories.

\subsection{Macro-stage: Coarse Prediction Stage}
\label{sec3.1.macro}
One of the most important factors in the uncertainty of the future behavior is the future heading of an individual. One way to narrow the possibilities is to be aware of the surroundings and learn patterns from the past. \cite{t++, AgentFormer, tnt} learn a representation of the environment, defined in image space, by encoding the semantic map of the scene into a 1D flattened feature, which can introduce distortion of spatial information of the scene.
% \comss{I don't think the 1-dimensionality is the issue here, but the compression, because 2D features can be encoded in 1D (e.g., ViT).}
For alignment between trajectories and the semantic map, trajectories $x$ are also represented in the pixel space as suggested in Y-net \cite{ynet}, using a Gaussian heatmap, denoted by $I_x$. 
The Gaussian filter has a variance of 4, and we create the homography matrices to map the world coordinates in meters to the image-based coordinates in pixel.
All past trajectories are represented in one heatmap centered at the last observed position, while the long-term and short-term goal positions are created per location.
% \comvp{I would describe briefly how the heatmap is constructed, if space allows it.} 
% \comml{Though I describe it here, figure 2 is explained in the next paragraph again.} \comvp{I meant to say to describe how a trajectory in the world coordinates is mapped to an image-based representation in the pixel coordinates. I.e., you need homography,  blurring kernel, etc.  The section below talks about local vs global view.}
Trajectories in $t_p$ past timestamp are all represented in a single heatmap, while each future step is represented as one heatmap per step. The trajectory heatmap size matches the size of the semantic map. % to be concatenated in the channel dimension.

Typically, the full environment information of a given scene is not necessary for long-term trajectory prediction. Often, the scene proximal to an agent's current location is sufficient. Thus, we focus only on the local semantic map, with trajectory heatmaps created as illustrated in \autoref{fig:local_map}. The local map is centered at the last observed agent location.  
The inputs and outputs of the Macro-stage are illustrated in \autoref{fig:macro_inout}. 
The input of the long-term goal prediction model, LG-CVAE, consists of concatenated (local semantic map, past trajectory heatmap) and outputs one long-term goal heatmap. The short-term goal prediction model, SG-net, has the input of concatenated (local semantic map, past trajectory heatmap, long-term goal heatmap) and outputs $N_{SG} + 1$ heatmaps, where  $N_{SG}$  is the number of short-term goals\footnote{ The extra count corresponds to the long-term goal.}. The local semantic map $I_M$ can be determined as $f(S,x_i^{t_p},\mathcal{H}, n)$ where $f$ is the function that converts the global scene information $S$ and homography $\mathcal{H}$ into a local image-based representation of size $(n,n )$ pixels centered at the last observed location $x_i^{t_p}$ of agent $i$. 

%\comml{do we explain how $n$ is decided here? It is based on per step pixel distance across both past and future trajectories, which is explained in the supplement.} \comsy{if explained in the supplementary, i think it's okay}

% \comvp{This section could use a few rounds of revisions to finetune the presentation.  I would actually encourage giving all elements (heatmaps, scene maps, etc) proper variable names. Also, it could be more formal. For instance, you could say that the local map image $M$ (or $I_M$, whatever notation we end up using) is computed as $M = f(S,x_i^{t_p},\mathcal{H},(w,h))$, where 'f' is the function that converts the global scene information and homography $\mathcal{H}$ into a local image-based representation of size $(w,h)$ pixels centered at the last-observed agent location $x_i^{t_p}$. }

% \noindent\textbf{LG-CVAE}\\
\vspace{-0.15in}
\subsubsection{LG-CVAE: Long-term Goal Prediction Model}
\vspace{-0.05in}
%\noindent\textbf{LG-CVAE: Long-term Goal Prediction Model.}
Where a person will go in the future depends primarily on the long-term goal position. Therefore, for different potential future trajectories, it is of paramount importance to predict different long-term goal positions in good quality.  
To model the inherent uncertainty with semantic map and heatmap trajectory representations, we combine U-net \cite{unet} and Conditional Variational AutoEncoder (CVAE) \cite{cvae} as studied in \cite{cvae-unet}. Given the heatmap $I_x$ of the past trajectories, the heatmap $I_{LG}$ of the long-term goal, and the local semantic map $I_M$, the objective of the CVAE is to maximize the conditional  distribution, 
% \comvp{I think the notation used here could be more intuitive.  For instance, I would use $I_M$ for the image-based representation of the local map, $I_x$ for the image-based representation of the trajectories, $I_{LG}$, etc.; $I_{something}$ immediately reminds the reader that this is in pixels.}
\small
\begin{equation}\label{eq:lgcvae}
p(I_{LG}|I_x,I_M) =  \int p_{\theta}(I_{LG}|w, I_x, I_M)p(w|I_x, I_M)dw.
\end{equation}
\normalsize
The stochasticity of the conditional latent distribution $p(w|I_x, I_M)$ is propagated and contributes to the multi-modality of $p(I_{LG}|I_x,I_M)$.
% By relieving the intractability \comvp{what does this mean?},
The LG-CVAE loss is defined as the negative evidence lower bound as follows.
\small
\begin{equation}\label{eq:lgcvae.elbo}
\begin{split}
\mathcal{L}_{I_{LG}} = - {\mathbb{E}}_{q_{\phi}(w|I_{LG}, I_x, I_M)}\left[{\log{p_{\theta}(I_{LG}|w, I_x, I_M)}} \right] \\
 + KL(q_{\phi}(w|I_{LG}, I_x, I_M)||p_{\psi}(w|I_x, I_M)),
\end{split}
\end{equation}
\normalsize
where $q_{\phi}(w|I_{LG}, I_x, I_M)$ and $p_{\psi}(w|I_x, I_M)$ are the posterior and the conditional prior distributions, respectively, assumed to be Gaussian for tractability. The output trajectory heatmap distribution $p_{\theta}(I_{LG}|w, I_x, I_M)$ has a Bernoulli distribution. Parameters of those densities are modeled using deep neural networks with the learning parameters $\phi$, $\psi$, and $\theta$, respectively, see \autoref{fig:arch}.
We use focal loss between the predicted heatmap $\widehat{I_{LG}}$ and the Ground Truth (GT) heatmap $I_{LG}$ for the reconstruction loss to mitigate the imbalanced class issue in the trajectory heatmap representation.

Joint pixel-based environment-trajectory input $(I_M, I_x)$ is encoded using a U-net architecture backbone \cite{unet}, which shows excellent performance on semantic segmentation learning.
% The encoded U-net features of dimension $(c, h, w)$, where $c$ denotes the channel dimension, $h$ and $w$ are the height and width of the feature map, are average-pooled in the spatial dimension, and outputs in $(c, 1, 1)$ feature maps, which is eventually converted to $c$-dimensional vector. 
The encoded U-net features of dimension $(C, H, W)$, where the feature map has $C$ channels, a height of $H$, and a width of $W$, are average-pooled in the spatial dimension, and outputs $(C, 1, 1)$ feature maps, which are eventually converted to a $C$-dimensional vector. 
% \comvp{should we say what the dimension of this feature is?  At least define it abstractly using a variable name, e.g., $D_{M,X}$?} 
It is concatenated with the latent factor $w$ sampled from the latent distribution. The posterior and the prior latent distributions are obtained from the separated posterior and prior network respectively consisting of convolutional layers.

To avoid the posterior collapse \cite{Bowman2015ALA, Snderby2016LadderVA} stemming from the strong U-net decoder, we pretrain the encoders and apply Free Bits \cite{Kingma2017ImprovedVI} and KL annealing \cite{Bowman2016GeneratingSF} strategies as studied in \cite{li2019emnlp}. Additional implementation details are discussed in the Supplementary Materials.

\vspace{-0.15in}
\subsubsection{SG-net: Short-term Goal Prediction Model}
\vspace{-0.05in}
%\noindent\textbf{SG-net: Short-term Goal Prediction Model}
In the second stage of the Macro-stage, we predict the short-term goals based on the long-term goal prediction from LG-CVAE. The purpose of SG-net is to give waypoints from the last observed step to the long-term goal that are well-aligned with the environment. 
The final stage in \autoref{sec3.2.micro} Micro-stage processes the trajectory and the semantic map as 1D feature vectors separately. Therefore, predicting all fine-grained future steps using only long-term goal information increases the risk of making predictions that are not well aligned with the environment based on destroyed spatial signals. SG-net utilizes U-net to generate $N_{SG} + 1$ heatmaps where  $N_{SG}$ is the number of short-term goals 
% \comvp{Why 'sh', is there any association between the acronym/var name and what it represents?  I would otherwise use something like $N_{SG}$ to denote the 'number of short term goals'.} 
and 1 accounts for the long-term goal as illustrated in \autoref{fig:macro_inout}. Unlike the LG-CVAE, this stage outputs the deterministic prediction based on the predicted long-term goal since we deal with the uncertainty of the fine trajectories other than long-term goals in the next stage. 
% \comvp{This modeling choice, ie deterministic is in contrast with the rest of the framework, so reviewers may focus on it unless it is justified.  Ie, why not a VAE here as well?}
Thus, SG-net loss is simply reconstruction loss with focal loss as follows.
\small
\begin{equation}\label{eq:sg.recon}
\begin{split}
\mathcal{L}_{SG} = - \sum_{i=1}^{N_{SG}+1} \big(
\alpha(1-\widehat{{I_{SG}}_i})^{\gamma} {I_{SG}}_i log(\widehat{{I_{SG}}_i}) \\
+ (1-\alpha)\widehat{{I_{SG}}_i}^{\gamma}(1-{I_{SG}}_i)log(1-\widehat{{I_{SG}}_i})
\big), 
\end{split}
\end{equation}
\normalsize
where ${I_{SG}}$ is the GT trajectory heatmap and $\widehat{{I_{SG}}}$ is the predicted heatmap and $\alpha=0.25, \gamma=2$ as studied in \cite{Lin2017FocalLF}.

\subsection{Micro-stage: Fine Prediction Stage}
\label{sec3.2.micro}
In the final stage of our model, we predict complete future trajectories at the micro level. 
Here we change the coordinate from the discrete pixel coordinate to continuous world coordinate for fine predictions.
Even if guided by predicted long-term and short-term goals from SG-net, individual steps may also have the variability stemming from the surrounding environment.
% \comvp{variability stemming from what?  better explain here or give examples of why it arises.} 
To deal with this uncertainty, we leverage CVAE in this step as well. As illustrated in \autoref{fig:arch}, we set $p(z|x)$ as the prior conditioned on past trajectories $x$, which is learned to approximate the posterior latent distribution $p(z|x, y)$ where $y$ denotes the future trajectories 
% \comvp{But you introduced different notation in first para of Sec.3; why not use that notation?}. 
In test time, we sample the latent factor $z$ from $p(z|x)$ to predict $p(y|z,x)$. While decoding future steps, our model use the long-term and short-term goal information from SG-net in the form of LSTM-encoded features.
We apply the Teacher Forcing technique to correct the prediction by feeding the GT/predicted long-term and short-term goals during training/test time respectively. 
To reduce the gap between training and test time reconstructions, we provide an additional reconstruction loss from the prior distribution following  \cite{Sohn2015LearningSO, cai2021syn}.
Thus, Micro-stage training loss with $\beta$-weighted ELBO \cite{Higgins2017betaVAELB} is formulated as follows.
\small
 \begin{equation}\label{eq:micro}
\begin{split}
&\mathcal{L}_{Micro} = - {\mathbb{E}}_{q_{\upsilon}(z|x, y)}\left[{\log{p_{\eta}(y|z, x)}} \right]  \\
&- {\mathbb{E}}_{p_{\tau}(z|x)}\left[{\log{p_{\eta}(y|z, x)}} \right]  
+ \beta KL(q_{\upsilon}(z|x, y)||p_{\tau}(z|x)),
\end{split}
\end{equation}
\normalsize
% where the first/second term is the reconstruction from the posterior/the prior distribution respectively, and the last term is the $beta$-weighted KL divergence between the posterior and the prior distribution. 
where both the latent distributions and the output trajectory distribution are assumed as Gaussian distributions.
We feed the U-net features from LG-CVAE to the prior network of Micro-stage so that the Micro-stage also recognizes the environment. Moreover, the Micro-stage encodes the GT/predicted $I_{LG}$ and $I_{SG}$s with a bi-directional LSTM and feeds them to the decoder in training/test time respectively, which helps the fine predictions of each step.
% \comvp{I find that this sections does not clearly describe how LG and SG are incorporated with the rest of the CVAE model.  For one, they are not a part of any of the loss terms, so that should be fixed.  You also simply say 'LG & SG' features are 'fed' into the GRU decoder, but any details are left out.  Also, Fig.3 illustrates that Unet features from LG-CVAE are conditioning z, but this is not described/justified/motivated. Most of the details will be in Supplement, and this should be stated, but some major points should be in this section to make it self-contained and complete. }
\section{Experiments}
\label{sec:5}
\vspace{-0.05in}
% We demonstrate the effectiveness of \MUSE on three datasets.
\autoref{sec:pre} introduces the datasets,  evaluation metrics, and statistical analysis used in the experiments.
\autoref{sec:quan} quantitatively evaluates SOTA models as well as \MUSE. \autoref{sec:qual} compares the qualitative aspects of the predictions for intuitive assessment.  % For a more intuitive assessment
In \autoref{sec:ablation}, each component of \MUSE is analyzed by ablation studies.% to determine its contribution.
% \comvp{needs a paragraph of introduction about the experiments/results you will be presenting}

\vspace{-0.05in}
\subsection{Preliminaries}
\label{sec:pre}
\noindent \textbf{Datasets } We used three datasets for the evaluation. The Stanford Drone Dataset (\textbf{SDD})~\cite{sdd} is used in the TrajNet challenge~\cite{sadeghian2018trajnet} and prior works~\cite{ynet, sophie}. The \textbf{nuScenes} Dataset~\cite{nusc} is a public autonomous driving dataset used by many prior arts~\cite{AgentFormer, PhanMinh2020CoverNetMB, Ma2020DiverseSF}. In addition, we created a new Path Finding Simulation Dataset (\textbf{PFSD}) using environments borrowed from~\cite{sohn2020laying}. Unlike SDD and nuScenes, the  spaces in these environments are significantly more complex to navigate. For more details, please refer to the Supplementary Materials.

%\commk{list and cite the various datasets, briefly describe the new synthetic dataset and refer to appendix for more details} 

\noindent \textbf{Evaluation Metrics } 
For the evaluation, we adopted the standard metrics of minimum Average Displacement Error (\textbf{ADE}) and Final Displacement Error (\textbf{FDE}).
%to be consistent with the prior work and perform fair comparisons. 
We also report the Kernel Density Estimate-based Negative Log Likelihood (\textbf{KDE NLL}) used in~\cite{trajectron, t++} as a comprehensive indicator of the predictive performance. Finally, we assess the Environment Collision-Free Likelihood (\textbf{ECFL})~\cite{a2x}, the probability that an agent has a path free of collision with the environment. We use it to address a drawback of existing works, which often neglect the importance of forecasting that adheres to environment structures.  We report ECFL in percent points, where 100\% means no collisions. More details can be found in the Supplementary Materials.

%\commk{list and cite the various metrics, justifying the need to study these metrics in combination and where prior work often cherry picks best results, refer to appendix for more details} 

\noindent \textbf{Statistical Analysis / Model Ranking } 
It is challenging to compare different models across multiple metrics. Therefore, we test the statistical significance of the results, using both traditional approach~\cite{JMLR:v7:demsar06a} and modern Bayesian analysis~\cite{JMLR:v18:16-305}. The Supplementary Materials provides the details.

%\commk{briefly describe that since it is challenging to compare models across multiple metrics, we perform a rank ordering statistical analysis, cite method, and refer to appendix for details. }

\vspace{-0.05in}
\subsection{Quantitative Results}
\label{sec:quan}
%To have thorough evaluation of the trajectory distribution, w
We conduct experiments on the three datasets introduced in \autoref{sec:pre} and compare the performance of \MUSE with Trajectron++ (T++) \cite{t++}, Y-net \cite{ynet}, and AgentFormer (AF) \cite{AgentFormer} baselines, using their public code. 
% For SDD results of Y-net and nuScenes results of AF, we use the provided pre-trained models. 
Scene maps  provided by PFSD and nuScenes show a much wider range of environments compared to SSD. Therefore, we provide a local view of the semantic map to all models including ours for a fair comparison.
% \comvp{I am not entirely sure what this means, can only guess. Can you clarify?}
For all experiments in \MUSE, we sample the latent factor $z$ only once in Micro-stage, and we gain all diversity from the latent factor $w$ in LG-CVAE by sampling it $K$ times since we assume the uncertainty primarily depends on the long-term goal position.

% \comvp{isn't this a common setting for all datasets?  If so, this should be specified in common experiment setup details.  I see this now almost repeatedly mentioned in each dataset.} 
% \comml{nuScenes has different setting.}
\autoref{tab:path} summarizes the experimental results on PFSD. Following the commonly used temporal horizon setting, we observe 3.2 sec (8 frames) and predict 4.8 sec (12 frames) future trajectories. 
Considering the increased complexity of the local environment layouts of PFSD, we choose sampling number $K=20, 50$ to investigate the learned trajectory distribution.
In ADE, our model can achieve the best performance in $K=50$ and the second-best in $K=20$. Although our model stands at second best in FDE, $K=20$, it leads in KDE NLL and ECFL performance. 
The KDE NLL scores of Y-net and AF indicate that their $K$ predictions fail to reflect the true trajectory distribution. This is because the $K$ predictions are indirectly sampled from the learned distribution from their first training stage but sampled in the next stage by manipulating them to focus on the diversity. Y-net conducts a test time sampling trick based on K-means clustering to obtain diverse predictions. AF has the second stage training to apply the pairwise distance loss between $K$ predictions for the diversity, which is inefficient since it requires re-training whenever $K$ changes. On the other hand, \MUSE can produce predictions within a low error range with GT trajectories, while reflecting the GT trajectory distribution (lower KDE NLL) and making realistic predictions reducing environment collisions (higher ECFL).
% \comvp{PFSD: in ADE, we either lead, or tie up with winners, in FDE we share 2nd place; but emphasize we win markedly in NLL and ECFL.}
% \comvp{SDD comparison missing: should emphasize that in ADE and NLL we do significantly better than competitors, while when we lose (FDE, ECFL), we largely tie up with the best method.  It would be good to comment why we don't lead in ECFL, when we do so consistently on the other two datasets.}

\autoref{tab:sdd} shows the evaluation on SDD. It follows the same temporal horizon setup as PFSD. As in the prior works, we choose $K=5, 20$ and errors are reported in pixel distance. \MUSE can significantly outperform the state-of-the-art methods in ADE. Though our model shows the second best performance in FDE, \MUSE largely ties up with the best method. For the same reason analyzed in PFSD, our model gives the best performance in KDE NLL. We can see that \MUSE has slightly worse ECFL, which is still the second best, than Y-net. This is because the labeling of the scene provided from Y-net is incomplete\footnote{The incomplete labels are discussed in the Supplementary Materials.}, which adversely affects \MUSE that relies heavily on the semantic map in Macro-stage predictions.

For the nuScenes dataset, following prior works, 2 sec (4 frames) observations and 6 sec (12 frames) predictions are made only for the vehicles and $K=5, 10$ generations are investigated. \autoref{tab:nu} shows that our model consistently outperforms the others in every metric and sampling number. Compared to the previous two datasets, nuScenes has a longer future time horizon and all agents are vehicles, which makes the prediction length of trajectories much longer. On the other hand, since nuScenes is a real world dataset, many static past trajectories are also observed. Due to the fact that our model focuses on learning the trajectory distribution rather than simply having min ADE/FDE based on diverse samplings and generations, these real world data characteristics in nuScenes are well reflected in the trained model, which can lead to better performance across all metrics.

% \comvp{We should add an overall assessment across the three datasets.  This is where computing average, across datasets, rank of each model could help or the Bayesian testing. In the absence of that evidence, we should emphasize the overall trend in leading or being 2nd best in ADE/FDE metrics, while strongly leading in NLL and ECFL.  Also attribute that to our modeling contributions, ie the inherent modeling of diversity (NLL) and the strong coupling between the env and the motion behavior, ie environment-aware, as reflected by ECFL. }

\begin{table}[t]
\caption{Results on the PFSD with $K$ = 20 and 50. With $t_p=3.2$s (8 frames) and $t_f=4.8$s (12 frames), errors are in meters.}
\label{tab:path}
\vspace{-1.5em}
\begin{center}
\scalebox{0.9}{
\begin{tabular}{c|c|c|c|c|c}
\toprule
$K$& Model & ADE $\downarrow$ & FDE $\downarrow$ & KDE NLL $\downarrow$ & ECFL $\uparrow$   \\
\midrule
% \midrule
% \multirow{4}{*}{5}  & T++
% &0.26 & 0.65& \textbf{0.57} & 83.21 \\
%                     & Y-net                   
% & 0.18 & \textcolor{blue}{0.34}& \textcolor{blue}{0.79} & 93.87\\
%                     & AF
% &\textbf{0.10} & \textbf{0.18 }& 1.33 &\textcolor{blue}{94.74} \\
%                     & Ours                    
% & \textcolor{blue}{0.15} & 0.37 & 0.89 & \textbf{97.44} \\
\midrule
\multirow{4}{*}{20} & T++            
&0.17 & 0.37& \textcolor{blue}{-0.88} & 83.32 \\
                    & Y-net                   
& 0.13 & 0.20 & 0.20 & 91.52 \\
                    & AF 
& \textbf{0.08} & \textbf{0.11} & 0.47 & \textcolor{blue}{94.54} \\
                    & Ours                    
& \textcolor{blue}{0.09} & \textcolor{blue}{0.19} & \textbf{-1.66} & \textbf{97.40} \\
\midrule
\multirow{4}{*}{50}  
& T++   
&0.14 & 0.25 & \textcolor{blue}{-1.11} &83.39\\
& Y-net   
&0.09 & \textcolor{blue}{0.12} & 0.04 & 91.74 \\
& AF   
&\textcolor{blue}{0.08} & \textbf{0.09} & 1.17 & \textcolor{blue}{95.37} \\
& Ours   
&\textbf{0.07} & 0.13 & \textbf{-1.94} & \textbf{97.50}\\
\bottomrule
\end{tabular}
}
\end{center}
\vspace{-1.5em}
\end{table}

\begin{table}[t]
\caption{Results on the SDD with $K$ = 5 and 20. With $t_p=3.2$s (8 frames) and $t_f=4.8$s (12 frames), errors are in pixels.}
\label{tab:sdd}
\vspace{-1.5em}
\begin{center}
\scalebox{0.9}{
\begin{tabular}{c|c|c|c|c|c}
\toprule
$K$& Model & ADE $\downarrow$ & FDE $\downarrow$ & KDE NLL $\downarrow$ & ECFL $\uparrow$     \\ % reporting 90.31
\midrule
\midrule
\multirow{4}{*}{5}  
                    & T++   
&\textcolor{blue}{11.11} & 24.42 & 8.74 & 86.94\\     
                    & Y-net      
& 11.49 & 20.19 & 8.98 & \textbf{89.99}\\
                    & AF   
&11.47 & \textbf{ 18.88 } & \textcolor{blue}{8.57} & 89.02 \\  
                    & Ours
& \textbf{9.60} & \textcolor{blue}{19.70}  & \textbf{8.43} &  \textcolor{blue}{89.30}\\
\midrule
\multirow{4}{*}{20}
                    & T++    
&8.16 &16.40 & \textcolor{blue}{7.37} & 86.88\\     
                    & Y-net   
& \textcolor{blue}{7.84}& 11.94& 8.05 & \textbf{89.32} \\
                    & AF 
& 8.35 &\textbf{ 11.03} & 7.48 & 87.30 \\       
                    & Ours                    
& \textbf{6.36} & \textcolor{blue}{11.10} & \textbf{7.21} &  \textcolor{blue}{89.30} \\       
\bottomrule

\end{tabular}}
\end{center}
\vspace{-1.5em}
\end{table}

\begin{table}[t]
\caption{Results on the nuScenes with $K$ = 5 and 10.  With $t_p=2$s (4 frames) and $t_f=6$s (12 frames), errors are in meters.}
\label{tab:nu}
\vspace{-1.5em}
\begin{center}
\scalebox{0.9}{
\begin{tabular}{c|c|c|c|c|c}
\toprule
$K$& Model & ADE $\downarrow$ & FDE $\downarrow$ & KDE NLL $\downarrow$ & ECFL $\uparrow$  \\
\midrule
\midrule
\multirow{4}{*}{5}  
& T++
&3.14 & 7.45 & \textcolor{blue}{7.20} & 68.99 \\

                    & Y-net                   
& 2.46 & 5.15 & 11.03 & 85.46\\
                    & AF  
&\textcolor{blue}{1.59}  & \textcolor{blue}{3.14} & 9.39 & \textcolor{blue}{86.74}\\
                    & Ours 
&\textbf{1.38} & \textbf{2.90} & \textbf{5.12} & \textbf{89.24} \\
\midrule
\multirow{4}{*}{10} 
& T++            
& 2.46 & 5.65 & \textcolor{blue}{5.61} & 69.02 \\
                    & Y-net                   
&1.88 & 3.47  & 7.52 & 82.90\\
                    & AF  
&\textcolor{blue}{1.30} & \textcolor{blue}{2.47} & 7.76 & \textcolor{blue}{85.76} \\
                    & Ours                    
& \textbf{1.09} & \textbf{2.10} & \textbf{3.82} & \textbf{89.33}\\       
\bottomrule

\end{tabular}}
\end{center}
\vspace{-2em}
\end{table}

%\commk{add a topic sentence here maybe?}}
\noindent \textbf{Statistical Analysis } We computed average rankings of the methods, and T++, Y-Net, AF, and \textbf{Ours} obtain 3.42, 2.92, 2.33, \textbf{1.33}, respectively. We conducted the Friedman test~\cite{10.2307/2279372} and confirmed that our method outperformed AF with statistical significance. We also conducted the Bayesian signed rank test~\cite{10.5555/3044805.3045007} and confirmed that our method is either superior or at least on par versus the competitors. The Supplementary Materials explain this in further detail.

\begin{figure*}[t]
\centering
\begin{subfigure}[b]{0.4\textwidth} 
{\includegraphics[width=\textwidth]{figures/legend.jpg}}
\end{subfigure}
\vspace{-2em}
\begin{flushright}

\begin{subfigure}[b]{0.29\textwidth} 
{\includegraphics[width=\textwidth]{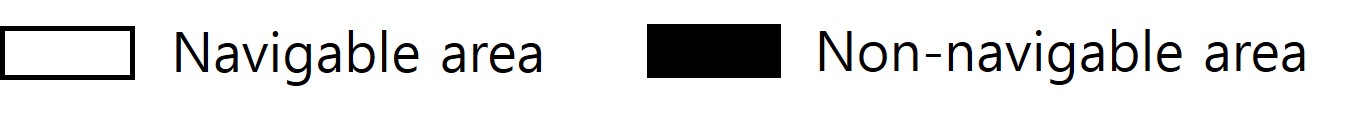}}
\vspace{-2.5em}
\end{subfigure}
\end{flushright}
\begin{subfigure}[b]{0.58\textwidth} 
{\includegraphics[width=\textwidth]{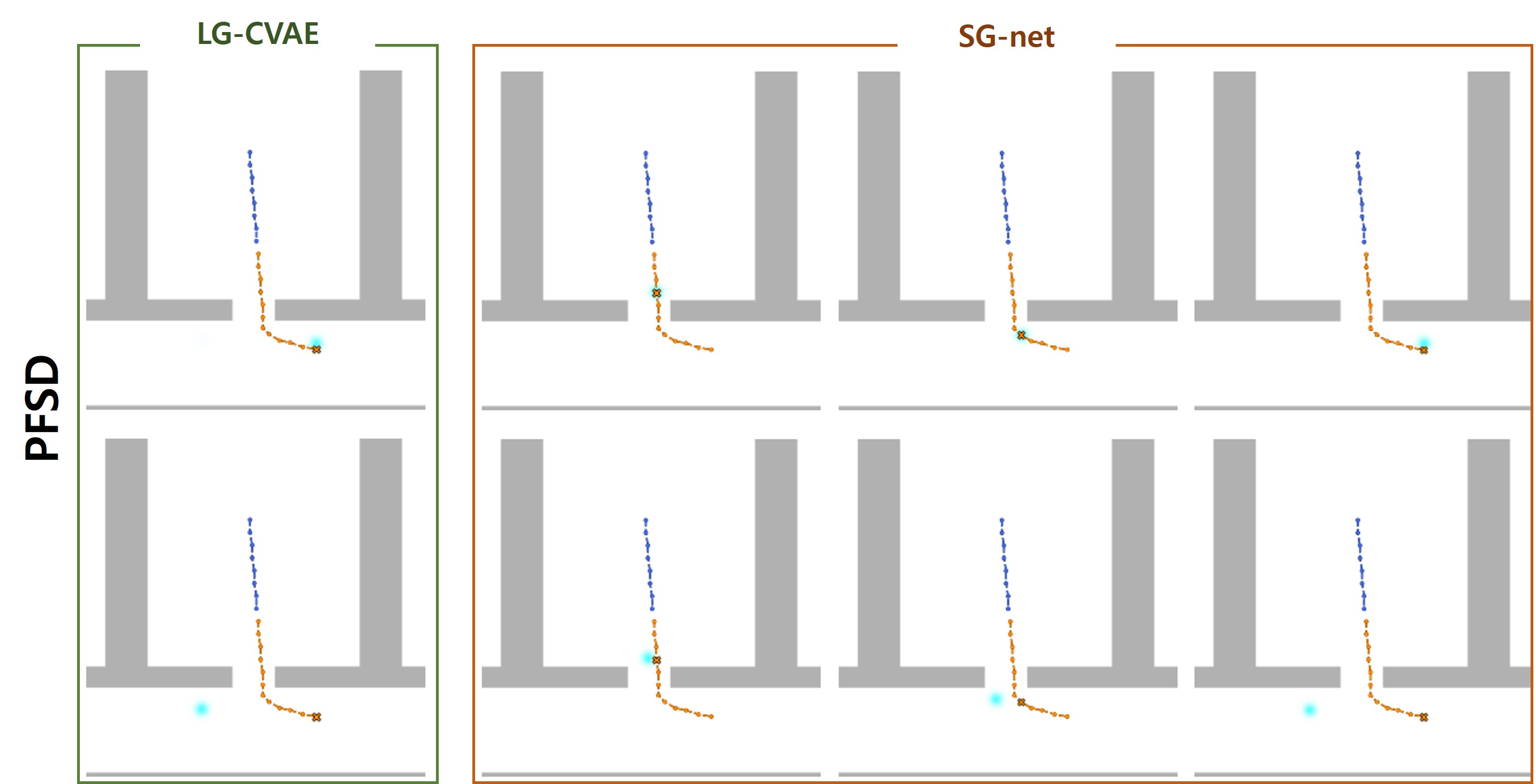} \caption{}\label{fig:macro_path} }
\end{subfigure}
\hspace{3em}
\begin{subfigure}[b]{0.27\textwidth} 
{\includegraphics[width=\textwidth]{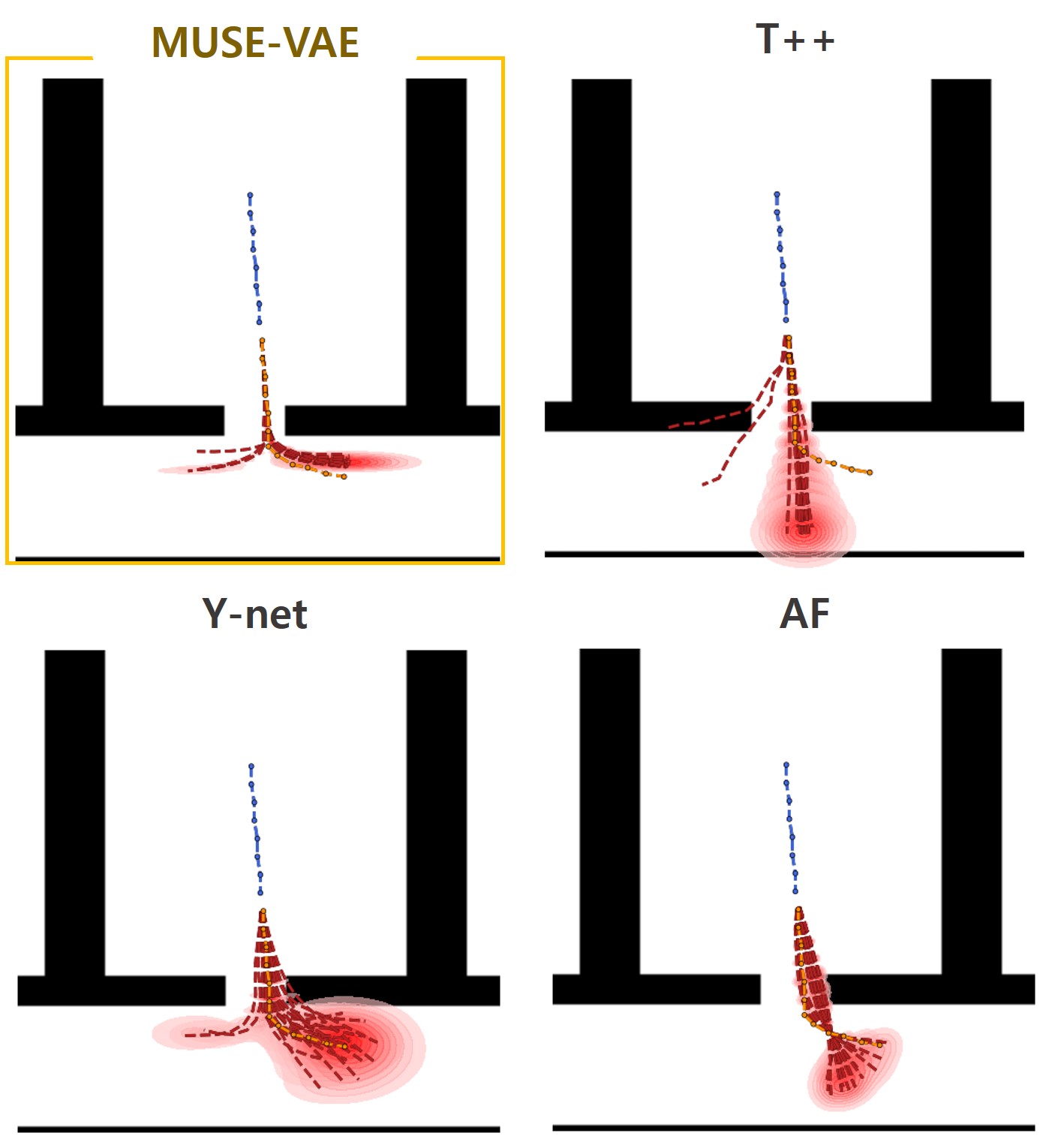} \caption{}\label{fig:micro_path} }
\end{subfigure}

\vspace{-1.5em}
\begin{flushright}
\begin{subfigure}[b]{0.28\textwidth} 
{\includegraphics[width=\textwidth]{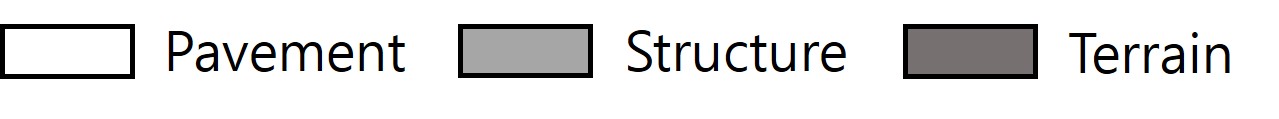}}
\vspace{-2.5em}
\end{subfigure}
\end{flushright}

\begin{subfigure}[b]{0.58\textwidth} 
{\includegraphics[width=\textwidth]{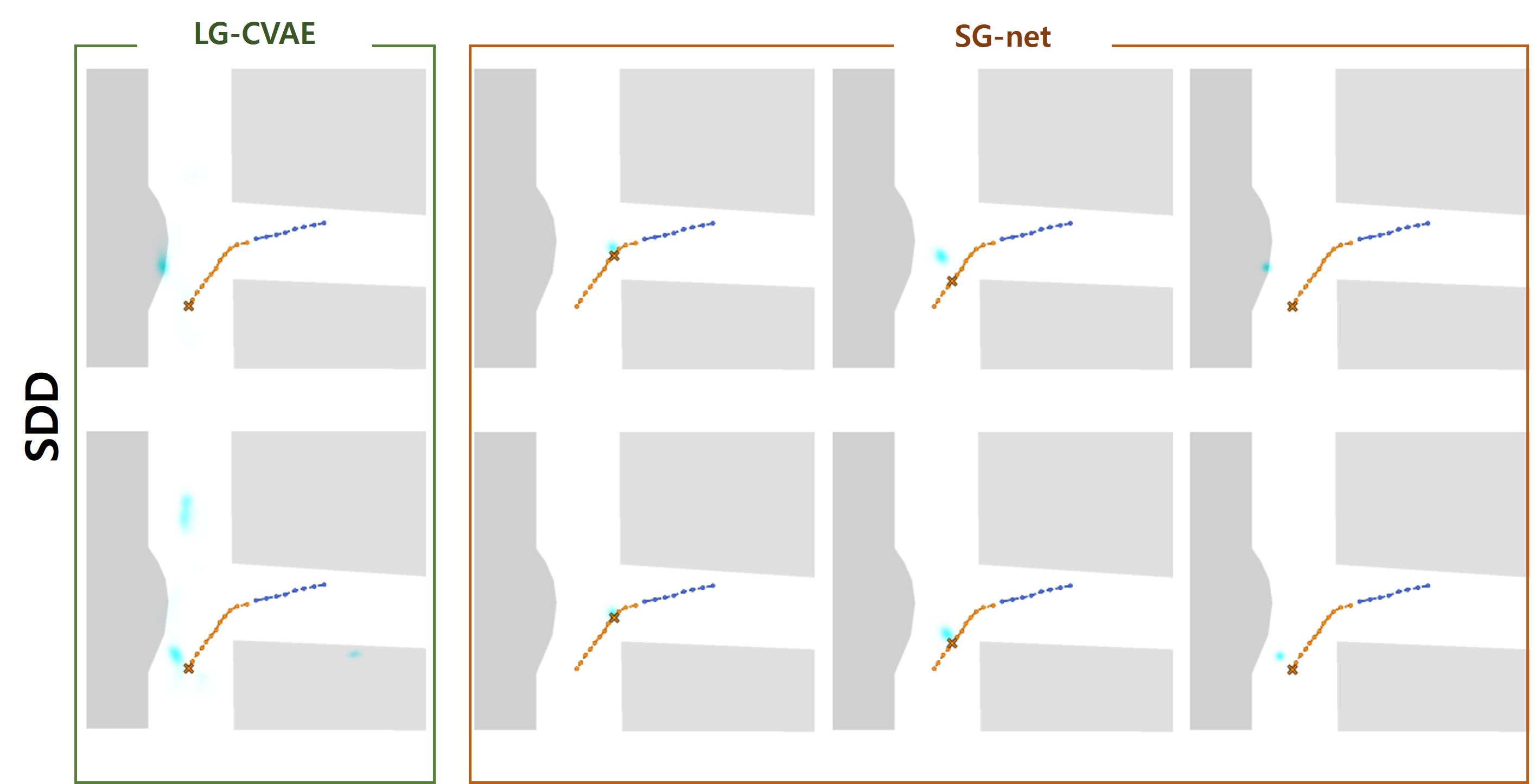} \caption{}\label{fig:macro_sdd} }
\end{subfigure}
\hspace{3em}
\begin{subfigure}[b]{0.27\textwidth} 
{\includegraphics[width=\textwidth]{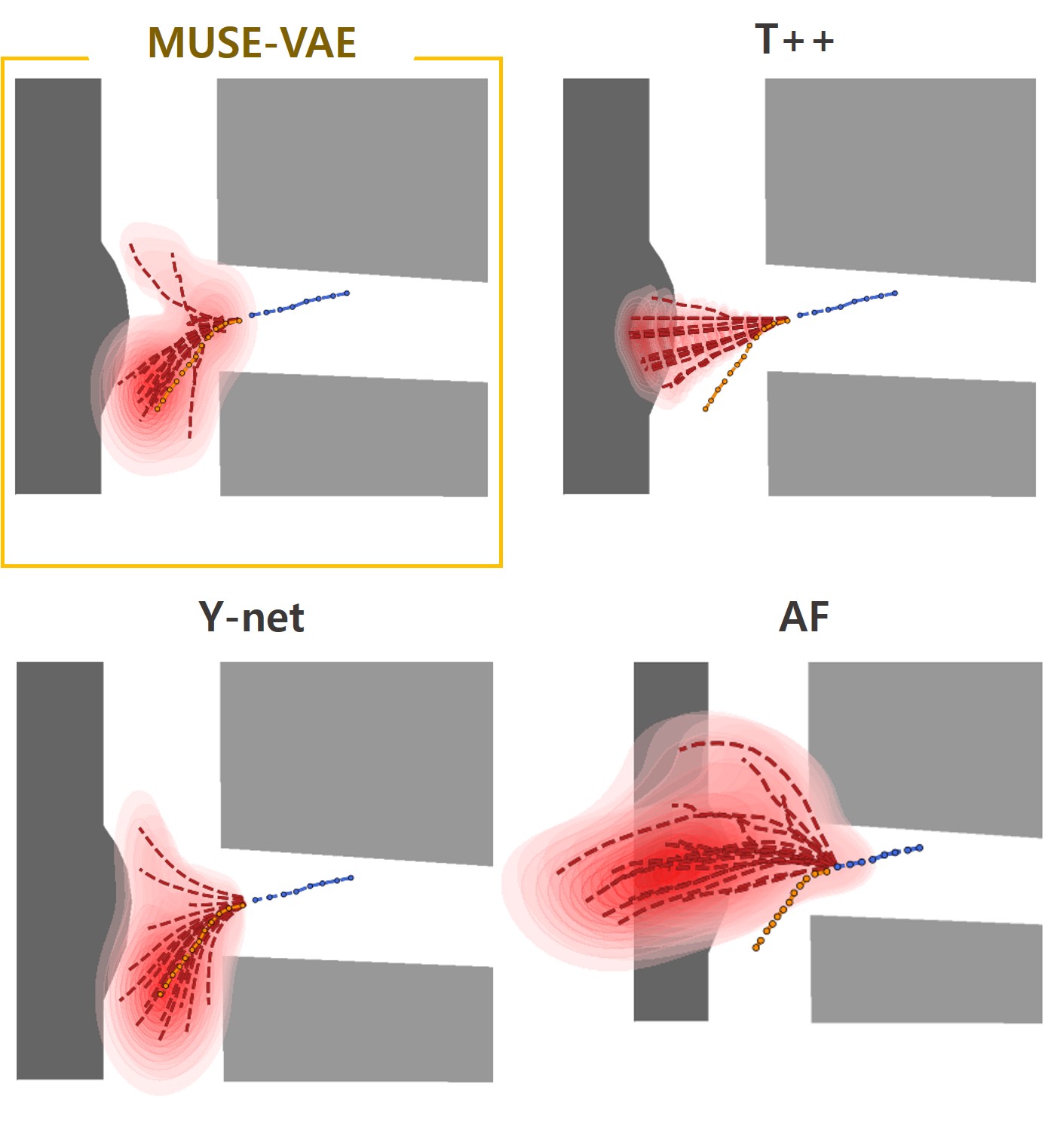} \caption{}\label{fig:micro_sdd} }
\end{subfigure}

\vspace{-1.5em}
\begin{flushright}
\begin{subfigure}[b]{0.5\textwidth} 
{\includegraphics[width=\textwidth]{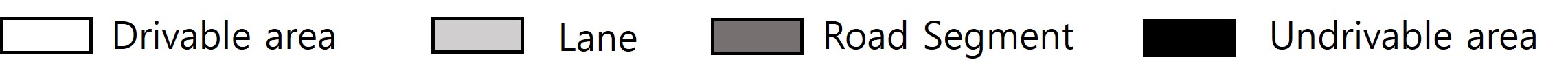}}
\vspace{-2.5em}
\end{subfigure}
\end{flushright}

\begin{subfigure}[b]{0.58\textwidth} 
{\includegraphics[width=\textwidth]{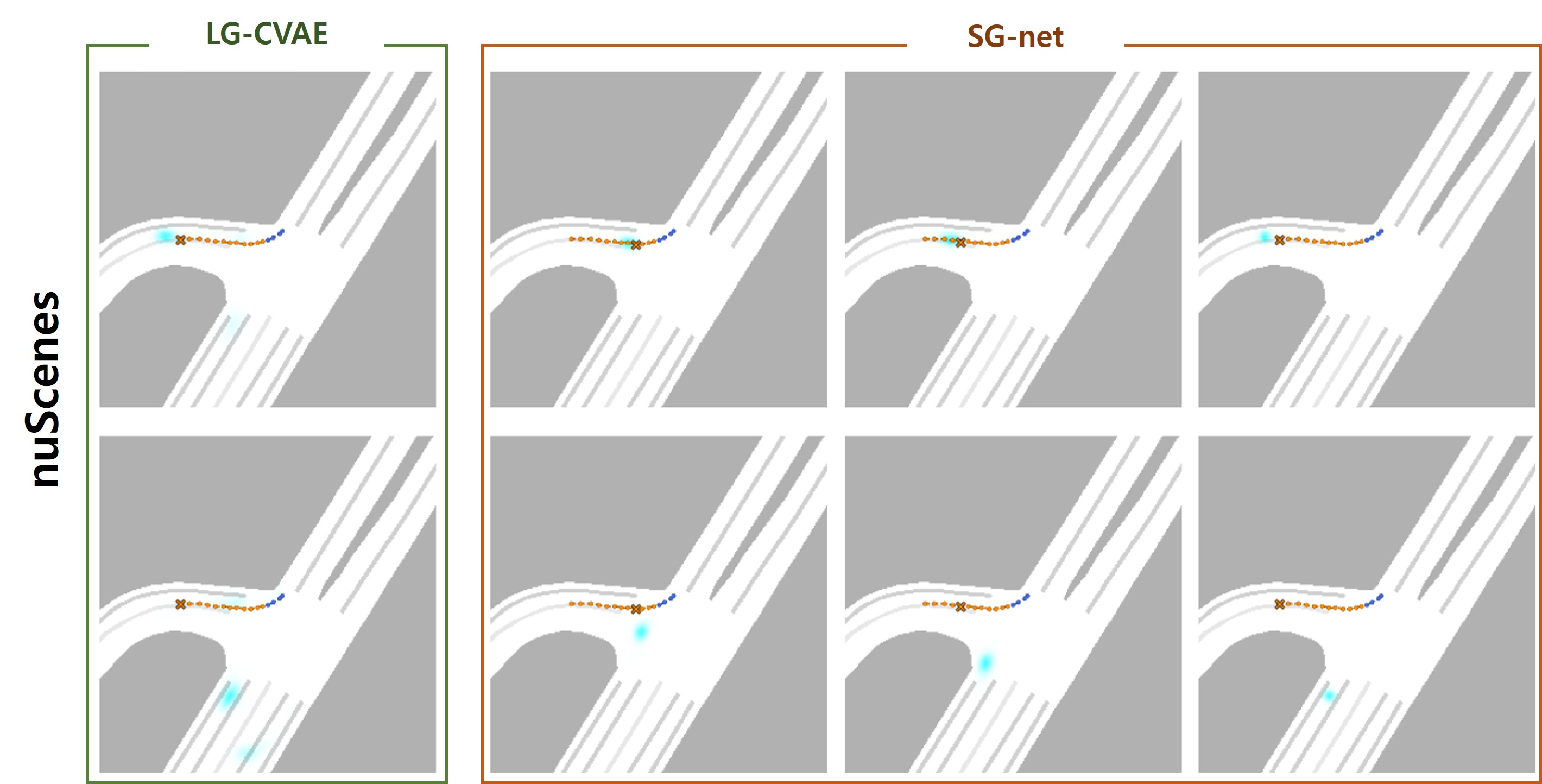} \caption{}\label{fig:macro_nu} }
\end{subfigure}
\hspace{3em}
\begin{subfigure}[b]{0.275\textwidth} 
{\includegraphics[width=\textwidth]{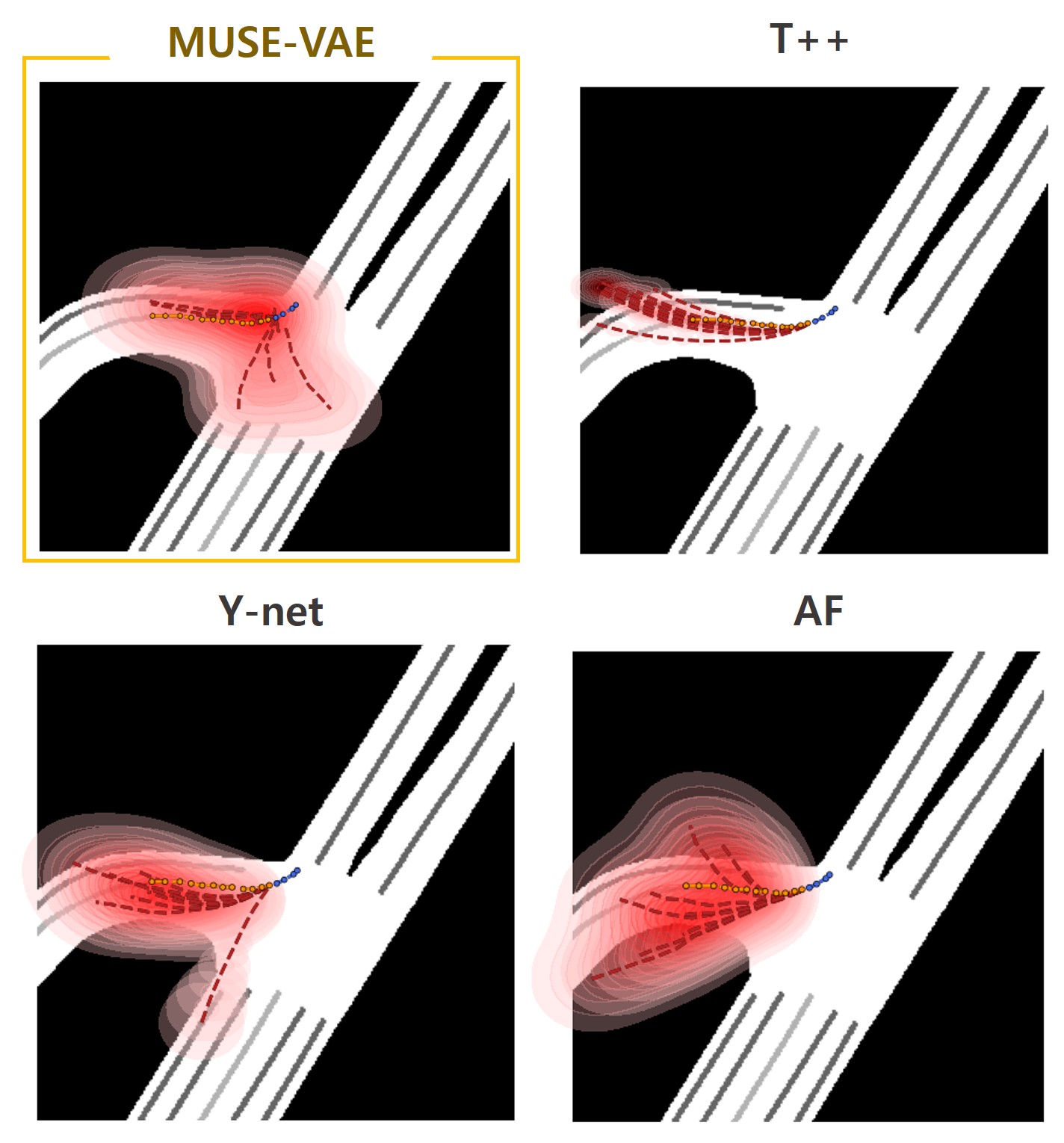} \caption{}\label{fig:micro_nu} }
\end{subfigure}

\caption{
Left: Macro-stage results of (a) PFSD, (c) SDD, and (e) nuScenes respectively. In the first column, the Long-term Goal (LG) heat map prediction from LG-CVAE is overlaid on the local semantic map. The following three columns are two Short-term Goals (SG) and one LG from SG-Net. Here we show only two different sampling generations in each dataset.  The blue and orange lines indicate GT past and GT future trajectories, respectively. GT LG and SGs are marked with `x’. 
Right: Complete trajectory predictions of (b) PFSD, (d) SDD, and (f) nuScenes respectively. In each dataset, the 1st/2nd/3rd/4th image from top-left to bottom-right is  from Micro-stage of ours/Trajectron++/Y-net/AgentFormer, respectively. The blue, orange, and red lines indicate GT past, GT future, predicted future trajectories, respectively.
}
\label{fig:plot}

\end{figure*}

\subsection{Qualitative Results}
\label{sec:qual}
We provide additional qualitative context to the quantitative metrics, in order to reveal the underlying factors that support each model's benefits and tradeoffs.
% To see intuitively how our model can produce realistic predictions, we further check the qualitative results in the followed section.
In \autoref{fig:plot}, we visualize several instances of  predicted long and short-term goals as well as the trajectories in the context of different environments and movement behaviors, driven by the three datasets we used for evaluation. Specifically, 
%\autoref{fig:macro_path} and \autoref{fig:micro_path} 
\cref{fig:macro_path,fig:micro_path} 
are instances from PFSD with $K=20$, %\autoref{fig:macro_sdd} and \autoref{fig:micro_sdd} 
\cref{fig:macro_sdd,fig:micro_sdd} 
are drawn for SDD with $K=20$, and 
%\autoref{fig:macro_nu} and \autoref{fig:micro_nu} 
\cref{fig:macro_nu,fig:micro_nu} 
come from nuScenes with $K=10$. 
We take a look at instances of a `fork-in-the-road' scenario from each dataset to test ability of models to understand the multimodality of long-term goals conditioned on the environment.
In \cref{fig:macro_path,fig:macro_sdd,fig:macro_nu}, we overlay predicted trajectories and goal heatmaps from Macro-stage over local semantic maps to demonstrate the ability of the models to make reasonable coarse predictions in the context of different environment features. The first column with the green border is the long-term goal prediction from LG-CVAE. The following three columns with the orange border are two short-term goals and one long-term goal from SG-Net.
% \comvp{are the LT goals from LG and SG guaranteed to be the same?  Or is the SG LT goal a refined version of the LT goal from LG?  If so, we should point to it and make an interesting story, if there is one.}. 
The two rows show two different predictions generated by sampling two different latent factors $w$ in LG-CVAE, based on the same observation $x$. We can see that (1) the short-term goals align well with the given predicted long-term goal, and (2) long-term goal projections naturally vary because of the structure of the `fork-in-the-road' scenario, which gives a generally bimodal uncertainty in the possible goal directions.
% \comvp{what would be nice, if possible for supplement, is to have instances of very similar env layouts, but modify them such that they suggest largely unimodal goal predictions.  Eg, in fig(a), the right-hand turn is blocked by a wall.  And see if the models are able to understand that and make unimodal goal predictions.}

%\autoref{fig:micro_path}, \autoref{fig:micro_sdd}, and \autoref{fig:micro_nu} 
\cref{fig:micro_path,fig:micro_sdd,fig:micro_nu} 
illustrate complete trajectory predictions, where the images in the clock-wise order, from the top-left, correspond to the  Micro-stage of \MUSE, followed by T++, AF, and Y-net, respectively.  Across all three datasets, we can observe that predictions of T++ and AF tend to lead to collisions with the environment. On the other hand, predictions of Y-net and our \MUSE are well-aligned and collision-free. We attribute this to T++ and AF encoding the semantic map into a 1D-representation, which entangles the spatial signal, while our model and Y-net process the semantic map along with the trajectory heatmap in 2D. 
Although Y-net produces predictions that avoid collision with obstacles, in contrast to \MUSE it yields trajectories with diverse duration, which often overshoot or undershoot the true trajectory horizon. This is because the goal predictions of Y-net are not made directly by the learned model; rather, they stem from the test time sampling trick, which is weakly conditioned on the past trajectory signal, particularly its velocity. On the other hand, our \MUSE's goal predictions are not only well aligned with the environment structure in the Macro-stage, but also reflect learned dependency on the past trajectory sequence modeled by an RNN in the Micro stage.
% \comvp{I really like this discussion!  The more discussions of this depth we and intuition we have, the stronger the paper will be.}

\begin{table}[t]
\caption{Ablation study on the PFSD with $K$ = 20. With $t_p=3.2$s (8 frames) and $t_f=4.8$s (12 frames), errors are in meters. 
}
\label{tab:ablation}
\vspace{-1.5em}
\begin{center}
\scalebox{0.88}{
\begin{tabular}{c|c|c|c|c}
\toprule
Model & ADE $\downarrow$ & FDE $\downarrow$ & KDE NLL $\downarrow$ & ECFL $\uparrow$  \\
\midrule
\midrule

\MUSE
& 0.09 & 0.19 & -1.66 & 97.40 \\ 
w/o SG-net                   
& 0.10 & 0.19 & -1.39 & 97.27 \\
w/o Micro-stage 
& 0.10  & 0.20 & - & 99.75 \\
w/o LL-prior
& 0.11 & 0.18 & -1.77 & 95.06 \\

\bottomrule

\end{tabular}}
\end{center}
\vspace{-2em}
\end{table}

\subsection{Ablation Study}
\label{sec:ablation}
We analyze the effectiveness of each component in \MUSE through an ablation study. \autoref{tab:ablation} shows three ablated experiments in addition to the complete model \MUSE. \textbf{w/o SG-net} model has no SG-net in Macro-stage, and thus, the long-term goal prediction is directly fed to the Micro-stage. \textbf{w/o Micro-stage} model does not include the Micro-stage, implying all future trajectories are predicted in the SG-net by letting $N_{SG}=t_f - 1$. In \textbf{w/o LL-prior} model, we eliminate the log-likelihood from the prior distribution $p_{\tau}(z|x)$ to assess the utility of this term in reducing the gap between the training and the inference-time reconstruction.

Our model requires the LG prediction produced by LG-CVAE, necessitating its presence in all experiments. LG-CVAE is also the primary factor influencing the variability of predictions. Thus, there is little observed variability in min ADE and min FDE. 
The most notable difference in performance stems from \textbf{w/o Micro-stage}, the absence of which precludes evaluation of the KDE NLL score. In this case,  complete trajectory predictions happen in the SG-net, defined in discrete pixel coordinates, thus limiting the accuracy of the forecasted trajectory\footnote{Complete trajectory predictions in this case are made from the heatmap maxima.}. 
%Thus, this pixel distortion makes it impossible to provide a KDE when the $K$ prediction is exactly the same. 
On the other hand, an advantage of this model is the few collisions, indicated by ECFL, because all predictions are obtained from pixel coordinates that are well aligned with the environment.
In \textbf{w/o SG-net}, Micro-stage has no information of waypoints other than the long-term goal predictions from LG-CVAE. Thus, the KDE NLL values shows that distribution learning of \textbf{w/o SG-net} is not as good as a complete model.
\textbf{w/o LL-prior} gives lower ECFL compared to other models. With reconstruction loss from the prior distribution during training, the model can learn how to produce predictions that better reflect the length of the past trajectory, which results in predictions not deviating too much from the given scene map. 
This thorough ablation study shows that it is crucial to consider both the Macro-stage for coarse predictions aligned well with the environment and the Micro-stage for fine predictions reflecting the past sequential states.

\section{Conclusion}
\label{sec:6}

% \commk{briefly restate what you did (can borrow from abstract) and summarize main results -- 1 para}
In this paper, we introduce \MUSE a probabilistic model capable of recognizing the environment and generating multimodal predictions based on the coarse-to-fine approach. 
Our experimental results using various datasets and metrics show \MUSE achieves both versatile and accurate forecasts that are well matched to environmental conditions.
% \noindent \textbf{Limitations and Future Work.} \commk{identify open challenges and future work opportunities}
\MUSE processes each agent independently, which cannot reflect agent-interaction. In the future work, we will take into consideration of multi agent-aware model that can avoid collisions with neighboring agents. 

%\commk{space permitting, try to have a more detailed discussion of future explorations }
%\comsy{need a separate heading...probably.} 

%\commk{general thumb rule to save space, avoid line spillovers in paragraphs/image captions by saving a few words from that paragraph. } % thank you!

\clearpage
% \comsy{we need a header for this appendix part "Supplementary Materials for ..."}
\appendix
\counterwithin{figure}{section}
\counterwithin{table}{section}
\setcounter{page}{1}

\begin{center}
\Large{
\textbf{Supplementary Materials for \MUSE}}
\end{center}

% \noindent This supplement consists of the following materials: 
% \begin{itemize}
% \item Datasets specification in \autoref{sec:dataset}
% \item Implementation details in \autoref{sec:impl}
% \item Explanation of the evaluation metrics in \autoref{sec:metric}
% \item Results of statistical validity test in \autoref{sec:test}
% \item Additional Visualization of experiments in \autoref{sec:vis}
% \item Limitation of SDD segmentation in \autoref{sec:sdd_seg}
% \item Challenges and Future Direction in \autoref{sec:challenges}
% \end{itemize}
In this supplement, we provide additional details about the proposed \MUSE, as well as the experimental evaluations, beyond those in the Main paper. \autoref{sec:dataset} offers dataset specifications for SDD, nuScenes, and PFSD, with scene examples of each dataset. \autoref{sec:impl} elaborates on the implementation details, including the model networks and the approach we used to create the local view of the semantic map. In \autoref{sec:metric}, we define the evaluation metrics used in the Main paper. \autoref{sec:test} presents details of two statistical significance tests, the Friedman test used in the Main paper, and the Bayesian Signed Rank test, whose results are shown here. Both tests offer additional evidence in support of improvements that \MUSE framework makes beyond the baseline models.  \autoref{sec:vis} supplement the qualitative analyses in the Main paper, showcasing instances of scenarios and the predictions made by all models in those scenarios, to highlight the different effects those models have on the forecasting process. \autoref{sec:sdd_seg} shows the limitation of the SDD segmentation provided by Y-net \cite{ynet} to explain the low ECFL discussed in \autoref{sec:quan} of the Main paper. 
% \comsy{briefly mention why this is something interesting and worths space} 
Finally, in \autoref{sec:challenges}, we discuss some key challenges of the trajectory prediction model and suggest possible directions for future research.

%\vspace{-0.5em}

\section{Datasets}
\label{sec:dataset}

\begin{figure}[b]
\vspace{-0.05in}
\centering
\begin{subfigure}[tbhp]{0.47\textwidth}
 {\includegraphics[width=\textwidth]{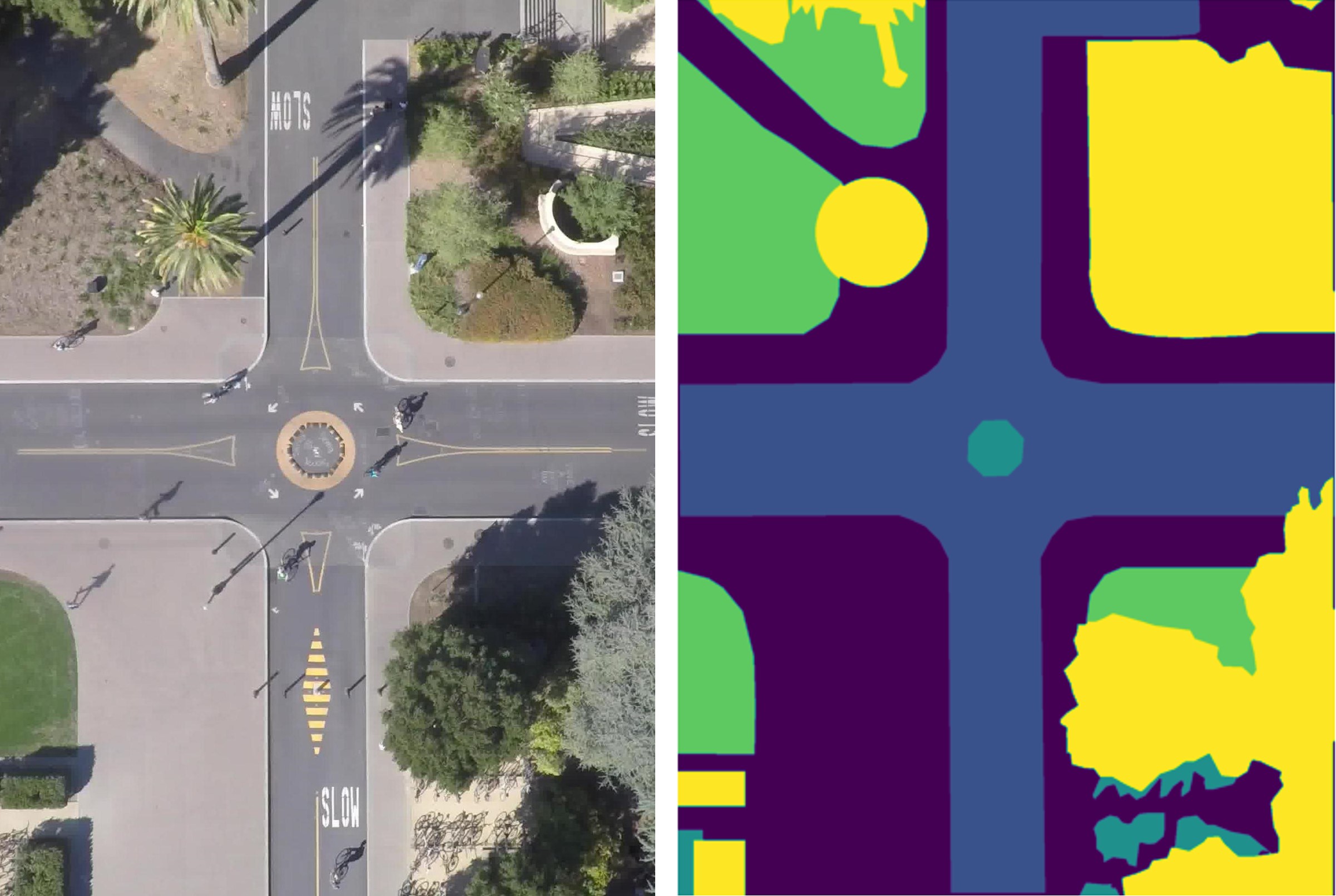} 
 \caption{SDD} \label{fig:sdd} 
 }
\end{subfigure}
% \begin{subfigure}[tbhp]{0.225\textwidth}
%  {\includegraphics[width=\textwidth]{figures/sdd_scene.jpg} 
%  \caption{SDD} \label{fig:sdd} 
%  }
% \end{subfigure}

\begin{subfigure}[tbhp]{0.232\textwidth}
 {\includegraphics[width=\textwidth]{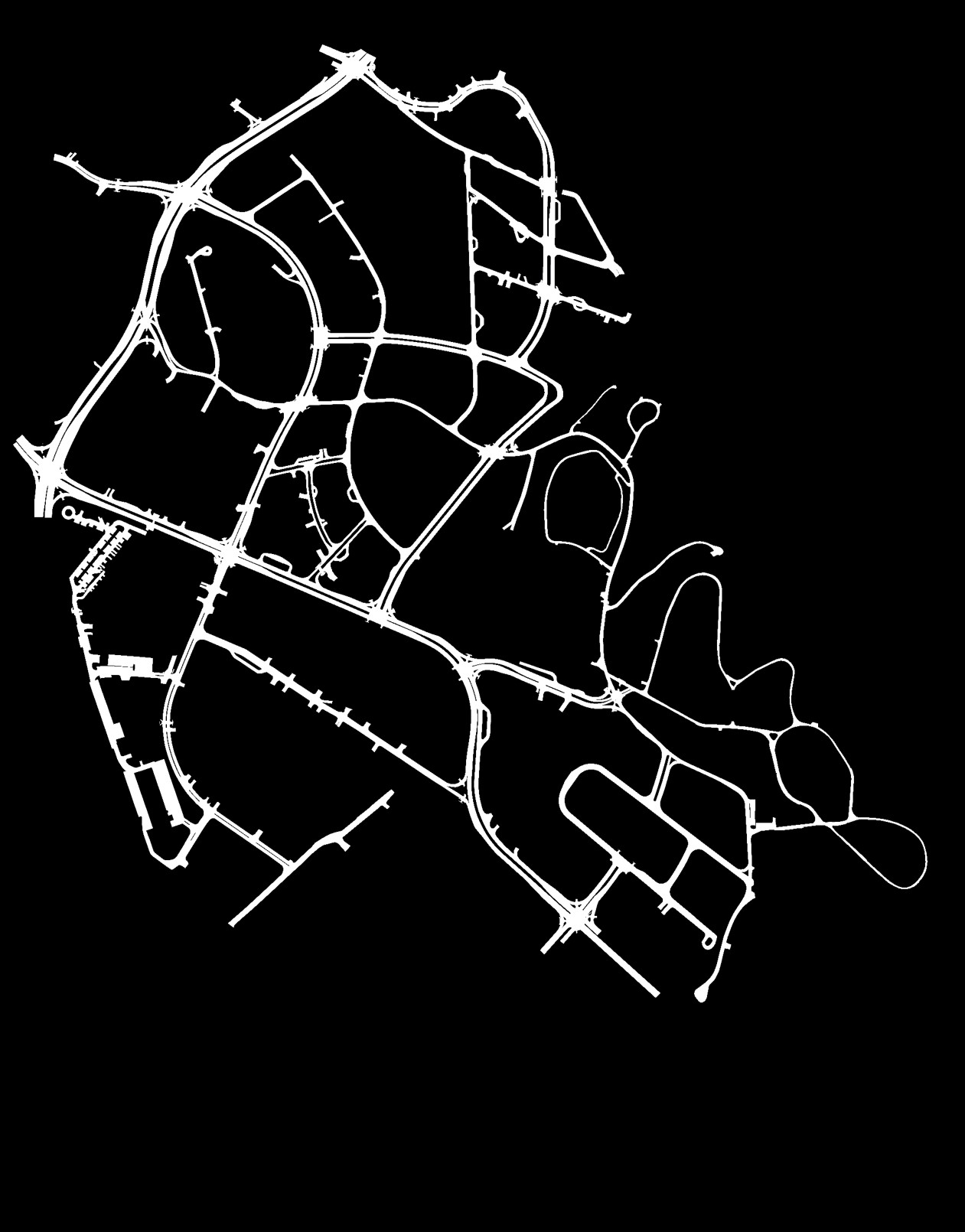} 
 \caption{nuScenes} \label{fig:nu} }
\end{subfigure}
\begin{subfigure}[tbhp]{0.23\textwidth}
 {\includegraphics[width=\textwidth]{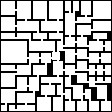} 
 \caption{PFSD} \label{fig:pfsd} }
\end{subfigure}

  \caption{(a) The global view of the scene image (left) and the semantic map (right).  The global view of the semantic map of (b) nuScenes, (c) PFSD.
}
\label{fig:datasets}
%\vspace{-1.5em}
\end{figure}

\subsection{Real World Datasets} 
The \textbf{Stanford Drone Dataset} (SDD)~\cite{sdd} consists of 20 unique scenes of college campus from bird-eye view collected by drones. It contains various agents such as pedestrians, cyclists, skateboarder, cart, car, and bus. We use the same split following the TrajNet challenge \cite{sadeghian2018trajnet}. As in \cite{ynet, sophie}, we sample at 2.5 Hz, which yields 3.2s (8 frames) observed trajectories and  4.8s (12 frames) future trajectories. We take advantage of the semantic map as well as the pixel data processed by \cite{ynet}. The semantic segmentation map is labeled as 5 classes; pavement, road, structure, terrain, and tree where each class has the class ID 1, 2, 3, 4, and 5, respectively. A sample scene image and its semantic map from SDD is shown in \autoref{fig:sdd}.

The \textbf{nuScenes Dataset}~\cite{nusc} is a public autonomous driving dataset. It provides 1,000 scenes in Boston, USA and Singapore and the corresponding HD semantic map with 11 annotated classes. Each scene is annotated at every 0.5s (2 Hz). Following the nuScenes prediction challenge setup, we split the train/val/test set, and predict only the vehicle category for 6s (12 frames) future trajectories based on 2s (4 frames) observations as in \cite{AgentFormer, PhanMinh2020CoverNetMB, Ma2020DiverseSF}.  \autoref{fig:nu} shows the global view of the binary map of the scene in Singapore with drivable (white-colored) area and undrivable (black-colored) area that nuScenes dataset provides.

\subsection{Synthetic Dataset} 
The \textbf{Path Finding Simulation Dataset} (PFSD) was generated by simulating the navigation of agents within 100 large synthetic environments borrowed from~\cite{sohn2020laying}.
These environments were designed according to the external shapes and interior organizations of rooms and corridors generally found in contemporary architecture~\cite{dogan2015optimization}.
Unlike SDD and nuScenes, the non-navigable spaces in these environments are significantly more complex for navigation.
Each of the environments was used to simulate 500 scenes (amounting to 50,000 total scenes), where a single agent navigates between two random points within the environment using the prevalent Social Force model~\cite{helbing1995social}.
As with SDD, the scenes were sampled at 2.5 Hz and further divided into training/val/test cases with 3.2s (8 frames) of observed trajectories and 4.8s (12 frames) of future trajectories.
We use subset of the PFSD and make the train/val/test set with 40/2/4 different synthetic environments, respectively. We provide an environment example in \autoref{fig:pfsd}. It is the binary map consisting of navigable (white-colored) and non-navigable (black-colored) space of the entire environment of one scenario. An agent finds a path by moving from a room to another room using the exit between obstacles. %Trajectories and maps of the five different environments are provided as the supplementary materials.

\section{Implementation Details}
\label{sec:impl}
\subsection{Local Semantic Map}
\noindent\textbf{Stanford Drone Dataset (SDD)} \\
We divide the semantic map class values (1 through 5) from Y-net \cite{ynet} with 5 so that the class values become 0.2 (pavement), 0.4 (road), 0.6 (structure), 0.8 (terrain), 1 (tree). 
We center the local view of the semantic map at the last observed step. As real-world agents have varying lengths of trajectories, for the radius of the local map we compute the per-step traversed distance of all trajectories, in each sequence (20 frames), and set the radius to be 20 times larger than the per-step distance.
Because the local semantic map is centered at the last observed position, it is possible that the local map region exceeds the original map. We represent those areas not in the original map as `non-navigable' space. We assume `structure' is the most non-navigable space among the five aforementioned classes, thus pad those areas with `structure' class value.
% \comsy{still confusing since this sounds like, in each map, the 'most' non-navigable, i.e. objects with least cross-over trajectories are different. is that what you meant? so it's possible that the outside can be padded with structures, neither trees nor roads?}
Each of the local map images and the Gaussian heatmaps for trajectories is resized into 256x256 pixels, then concatenated in the channel dimension.

\noindent\textbf{nuScenes Dataset} \\
We use the official code of AgentFormer~\cite{AgentFormer} to preprocess the nuScenes semantic map. This results in a 3 channel semantic map with four categories: drivable area, lane, road segmentation, and undrivable area. We further preprocess this information to create a single-channel semantic map by setting the drivable area, lane, road segmentation, and undrivable area as 0, 0.3, 0.6, and 1, respectively.
To determine a local map size, we use the same policy as in SDD. 
For the local map region out of the original map, we pad it with the `undrivable area' class value.

\noindent\textbf{Path Finding Simulation Dataset (PFSD)} \\
Since the synthetic dataset has consistent step size throughout the data, we compute the average per-step distance across the entire training set, about 8-pixel distance. Based on this, the local view of the semantic map is centered at the last observed position of the agent and its size is $160\times160$. We encode the navigable / non-navigable space as the values 0 / 1, respectively. The areas of the local map that deviate from the original map are padded by value 1 to indicate non-navigable space.

\subsection{Networks}
We implement \MUSE in PyTorch. All networks are trained with Adam optimizer~\cite{DBLP:journals/corr/KingmaB14}.
LG-CVAE has the backbone of U-net~\cite{unet} combined with CVAE. U-net encoder blocks consist of [32, 32, 64, 64, 64] output channel dimensions with the input channel 2  consisting of a local map and a heatmap for past trajectories. The decoder blocks have [64, 64, 64, 32, 32] output channel dimensions with the final output channel 1 to predict the long-term goal heatmap. The posterior network consists of convolutional layers with same output channels as the U-net encoder blocks. The prior network takes the feature from the U-net encoder and process it further using two convolutional layers with output channel dimension [32, 32]. Following \cite{cvae-unet}, the resulting 2D feature map is average-pooled into 1x1, then fed to a 1x1 convolutional layer to estimate the mean and the standard deviation of the posterior and the prior latent distribution, with the dimension set to 10.
To avoid the posterior collapse, the encoder of LG-CVAE is pretrained with the AE loss for 10 epochs with the learning rate of $1e^{-3}$. During training of LG-CVAE with VAE loss, we anneal the KL loss for the first 10 epochs; FB is set as 0.8, 6, and 3 for PFSD, SDD, and nuScenes, respectively. The learning rate is $1e^{-3}, 1e^{-4},$ and $1e^{-4}$ for PFSD, SDD, and nuScenes, respectively.

SG-net is also based on the U-net. It has one additional block of 128 output channel dimensions more than LG-CVAE. The input channel of the encoder is 3, for a local map, a heatmap for past trajectories, and a heatmap for a long-term goal. The final output channel of the decoder is $N_{SG}+1$ for $N_{SG}$ heatmaps of $N_{SG}$ short-term goals and a heatmap of a long-term goal.
The learning rate is $1e^{-3}, 1e^{-4},$ and $1e^{-3}$ for PFSD, SDD, and nuScenes, respectively.

In Micro-net, we utilize the position, velocity, and acceleration of the past sequence as in \cite{t++}. The prior network consists of an LSTM with 64 hidden dimensions and 2 FC layers with the output dimensions [256, 40] to estimate the mean (20D) and the standard deviation (20D) of the prior latent distribution. The 256 dimensional hidden feature from the prior network is processed once more by concatenating it with the feature from the LG-CVAE, which encodes the semantic map using FC layer with 32 output dimensions in order to give the map information to the decoder. The posterior network consists of a bi-directional LSTM with 64 hidden dimensions, followed by two FC layers with [256, 40] to estimate the mean (20D) and the standard deviation (20D) of the posterior latent distribution. 
The decoder has a GRU with 128 hidden dimensions, followed by FC layers to predict the mean and standard deviation of the 2D position distribution. The short-term goal heatmap predictions from Macro-stage are converted to the 2D position and encoded by bi-directional LSTM with 64 hidden dimensions and further processed into a 2D feature by FC layer and fed to the GRU. 
We use the learning rate $1e^{-3}$ and $\beta=50$ for all datasets. FB is 0.07 for PFSD and nuScenes, and 1 for SDD.

A subset of our code for PFSD is provided as additional supplementary material.%\comvp{do not forget to do this!}. 
The complete code for \MUSE will be released upon acceptance, following the conference policy.

\section{Evaluation metrics}
\label{sec:metric}

To evaluate the performance, we use four metrics: Average Displacement Error (ADE), Final Displacement Error (FDE), Kernel Density Estimate-based Negative Log Likelihood (KDE NLL), and Environment Collision-Free Likelihood (ECFL).

\noindent\textbf{Average Displacement Error (ADE)} Given $t_f$ future timestamps, ADE is defined as the $L_2$ distance between the future GT and predictions which is averaged over $t_f$. Following prior works \cite{socialgan, sociallstm, trajectron, ynet, AgentFormer, t++}, we report the minimum ADE among $K$ ADEs obtained from $K$ predictions. 
% Here we report average and standard deviation \comsy{this is no longer true} as well as minimum of the $K$ predictions for better understanding of the predicted distribution.

\noindent\textbf{Final Displacement Error (FDE)} FDE is $L_2$ distance between the GT and prediction at the final future step $t_{p+f}$. Same as ADE, the minimum FDE among $K$ predictions is reported.
% investigate the average and standard deviation of the $K$ FDEs. \comsy{this is no longer true}

\noindent\textbf{Kernel Density Estimate-based Negative Log Likelihood (KDE NLL)} To determine if the generative model learns the characteristics such as variance and multi-modality of the distribution, \cite{trajectron, t++} introduce KDE NLL. First, the pdf is estimated by Kernel Density Estimate (KDE) using the $K$ sampled predictions at each future timestep, and then the mean log-likelihood of the GT trajectory is obtained based on the pdf. We adopt the approach in \cite{t++} and their publicly released code.

\noindent\textbf{Environment Collision-Free Likelihood (ECFL)} Realistic trajectory predictions should not violate environmental restrictions. \cite{a2x} proposes ECFL, the probability an agent has a path that is free of collision with the environment defined as 
    $ ECFL(p, E) = \frac{1}{k}\sum_{i=1}^{k}{
        \prod_{t=1}^{t_{f}}{
            E[p_{i,t,0}, p_{i,t,1}]
        }
    },$
where $E$ is the scene environment represented as a binary map with 1s and 0s indicating the navigable and the non-navigable spaces, respectively. $p$ are the $K$ predicted positions of an agent under the temporal horizon $t_f$. We report ECFL in percent points, where 100\% means no collisions.

% ----------------------------------------------------------------------------------------

\section{Statistical Validity Test}
\label{sec:test}
Since our evaluation used multiple datasets and four measures, we conducted additional analyses using the average rank~\cite{JMLR:v7:demsar06a} and the Bayesian statistical validity analysis~\cite{JMLR:v18:16-305} to assess the significance of the obtained results.

\subsection{The Friedman Test}\label{sec:friendman}
We borrow notations from~\cite{JMLR:v7:demsar06a} in this section. We first calculated the Friedman statistic~\cite{10.2307/2279372} as
\begin{align}
    \chi_{F}^{2} &= \frac{12N}{k(k+1)} \left[ \sum_{j} R_{j}^{2} - \frac{k(k+1)^{2}}{4} \right],
\end{align}
where we compare $k$ methods tested on $N$ datasets. Here, $R_{j}$ denotes the average ranks of algorithm $j$ over all $N$ datasets, i.e.,
\begin{align}
    R_{j} &= \frac{1}{N} \sum_{i} r_{i}^{j},
\end{align}
where $r_{i}^{j}$ denotes the rank of $j$-th method among $k$ algorithms tested on $i$-th dataset among $N$ total datasets. One can approximate the probability distribution of the value as a Chi-square distribution. If $k$ or $N$ is small, one needs to find exact critical value from the precomputed table. Iman and Davenport~\cite{doi:10.1080/03610928008827904} proposed a better statistic using the $\chi_{F}^{2}$,
\begin{align}
    F_{F} &= \frac{(N-1) \chi_{F}^{2}}{N(k-1) - \chi_{F}^{2}}
\end{align}
and this follows the F-distribution with $(k-1)$ and $(k-1)(N-1)$ degrees of freedom.

In our case, we have four methods to compare ($k = 4$) and 24 datasets ($N = 24$). We considered each evaluation setting (hyperparameter ($K$) choices, datasets, performance measures) as different datasets ($2 \times 3 \times 4$). Henceforth, we look for the F-distribution's critical value for 3 and 69 degrees of freedom. At 95\% confidence level, the (upper) critical value is 2.737. Using the ranks obtained from our quantitative result, $F_{F}$ is 21.278 which is significantly larger than the critical value, \textbf{rejecting the null hypothesis}, which states that all methods are equivalent. 

As a post-hoc test, we conducted the Nemenyi test~\cite{Nemenyi1963}. In the test, if two methods' average rank difference is larger than the critical difference defined as
\begin{align}
    CD &= q_{\alpha} \sqrt{\frac{k(k+1)}{6N}},
\end{align}
then there is a significant performance difference between the two methods. Here, $q_{\alpha}$ is 2.569 for $k = 4$ at 95\% confidence level, hence $CD = 0.957$. Since the average rank of our method is 1.33 and that of AF is 2.33, we argue that \textbf{our method outperformed AF} in the evaluation. Note that the average ranks of Y-Net and T++ are 2.92 and 3.42, respectively.

\subsection{Bayesian Signed Rank Test}\label{sec:bsrt}

We also provide significance testing result based on modern Bayesian statistical validity analysis to address potential limitations of the traditional frequentist null hypothesis significance testing~\cite{JMLR:v18:16-305}. We ran the Bayesian signed-rank test~\cite{10.5555/3044805.3045007} for each pair of methods and for each measure. This test also accounts for the region of practical equivalence (ROPE)~\cite{Kruschke2018}. If the difference between two methods is smaller than the ROPE, then there is no practical difference in performance. 

In our evaluation, we have several metrics and datasets and each needs a careful definition of ROPE to conduct a proper analysis. First, for ADE and FDE, we adopted the standard 0.5 meter difference as the ROPE~\cite{t++}. However, one of the datasets we used, SDD, does not have the geometric calibration data to obtain metered measures in its test set, unlike PFSD and nuScenes. Henceforth, prior works, e.g.,~\cite{ynet}, used pixel differences to calculate the ADE and FDE. Therefore, we used 1 pixel difference for the ROPE, considering the resolution of the image and approximate sizes of real world structures and objects in the scene. It should be noted that, we also tested with a larger ROPE (3 pixels), but there was no change in the conclusion of this analysis. For KDE NLL, it is challenging to define ROPE since NLL is not a scale, but a likelihood value. So we set ROPE as zero for NLL. For ECFL, since it has same scale as accuracy $[0, 100]$, we use the standard 1\% difference for the ROPE.

In \cref{tab:baycomp1,tab:baycomp2,tab:baycomp3,tab:baycomp4}, we report the Bayesian signed-rank pairwise test result for $C(4,2)=6$ comparisons. 
%The corresponding win/tie/loss scores and overall ranks for each metric can be seen in \autoref{tab:avgrank}. 
\autoref{tab:avgrank} summarizes all the aforementioned pairwise results by computing the  average ranks of each method, in each of the tables \cref{tab:baycomp1,tab:baycomp2,tab:baycomp3,tab:baycomp4}, based on the number of times the method "won", "tied", or "lost" in the pairwise comparison.  For instance, in \autoref{tab:baycomp3}, \MUSE won 3-out-of-6 times, T++ 2/6, AF 1/6, and Y-Net 0/6 times, resulting in ranks of 1, 2, 3, and 4 for the four methods, respectively. Based on this, and in line with the traditional frequentist analysis in \autoref{sec:friendman}, we conclude that 
\textbf{our \MUSE outperforms the SOTA competitors}, on average, across all datasets and measures.
%\comvp{Where did the calculations re average rank, as I did in my comments in the pdf I sent, go?  You should include the average ranks of all methods (the one last table I had), which summarize the pairwise comparisons in an easy-for-reader-to-digest format.  Else, one is forcing the reader to figure this out from Tab D1-D4 and potentially come up with an alternate conclusion.}

\begin{table}[tbp]
\centering
\caption{Comparing ADE of methods using Bayesian signed-rank test. For PFSD and nuScenes, ROPE is defined as 0.5 meters. For SDD, ROPE is 1 pixel.}
\label{tab:baycomp1}
\scalebox{0.9}{
\begin{tabular}{c|c|c|c|c}
\toprule 
\multicolumn{5}{c}{\textbf{PFSD, nuScenes}}     \\
\midrule
% \multicolumn{5}{c}{Bayesian signed-rank test}     \\
% \midrule
Method A & p(A $>$ B)  & p(A $\approx$ B)  & p(A $<$ B)  & Method B \\
\midrule
T++     & 0.00     & \textbf{0.87}    & 0.13     & Y-Net       \\
T++     & 0.00     & 0.27    & \textbf{0.73}     & AF       \\
T++     & 0.00     & 0.27    & \textbf{0.73}     & Ours       \\
Y-Net   & 0.00     & \textbf{0.87}    & 0.13     & AF       \\
Y-Net   & 0.00     & \textbf{0.59}    & 0.41     & Ours       \\
AF      & 0.00     & \textbf{1.00}    & 0.00     & Ours       \\
% \midrule
% \multicolumn{5}{c}{Score}     \\
% \midrule
% Method & \# win & \# tie & \# lose & Rank \\
% \midrule
% T++    & 0      & 0      & 3       & 4    \\
% Y-net  & 1      & 0      & 2       & 3    \\
% AF     & 2      & 1      & 0       & 1.5  \\
% Ours   & 2      & 1      & 0       & 1.5 \\
\bottomrule
\end{tabular}
}

% \vspace{0.5em}
% \begin{tabular}{ccccc}
% \toprule 
% \multicolumn{5}{c}{PFSD, nuScenes}        \\
% \midrule
% Method & \# win & \# tie & \# lose & Rank \\
% \midrule
% T++    & 0      & 0      & 3       & 4    \\
% Y-net  & 1      & 0      & 2       & 3    \\
% AF     & 2      & 1      & 0       & 1.5  \\
% Ours   & 2      & 1      & 0       & 1.5 \\
% \bottomrule
% \end{tabular}

\vspace{1em}
\scalebox{0.9}{
\begin{tabular}{c|c|c|c|c}
\toprule
\multicolumn{5}{c}{\textbf{SDD}}     \\
\midrule
% \multicolumn{5}{c}{Bayesian signed-rank test}        \\
% \midrule
Method A & p(A $>$ B)  & p(A $\approx$ B)  & p(A $<$ B)  & Method B \\
\midrule
T++     & 0.00     & \textbf{1.00}    & 0.00     & Y-Net       \\
T++     & 0.00     & \textbf{1.00}    & 0.00     & AF       \\
T++     & 0.00     & 0.27    & \textbf{0.73}     & Ours       \\
Y-Net   & 0.00     & \textbf{1.00}    & 0.00     & AF       \\
Y-Net   & 0.00     & 0.27    & \textbf{0.73}     & Ours       \\
AF      & 0.00     & 0.27    & \textbf{0.73}     & Ours       \\
% \midrule
% \multicolumn{5}{c}{Score}        \\
% \midrule
% Method & \# win & \# tie & \# lose & Rank \\
% \midrule
% T++    & 0      & 2      & 1       & 3.5  \\
% Y-net  & 0      & 3      & 0       & 2    \\
% AF     & 0      & 2      & 1       & 3.5  \\
% Ours   & 2      & 1      & 0       & 1   \\
\bottomrule
\end{tabular}
}

% \vspace{0.5em}
% \begin{tabular}{ccccc}
% \toprule
% \multicolumn{5}{c}{SDD}        \\
% \midrule
% Method & \# win & \# tie & \# lose & Rank \\
% \midrule
% T++    & 0      & 2      & 1       & 3.5  \\
% Y-net  & 0      & 3      & 0       & 2    \\
% AF     & 0      & 2      & 1       & 3.5  \\
% Ours   & 2      & 1      & 0       & 1   \\
% \bottomrule
% \end{tabular}
\end{table}

\begin{table}[tbp]
\caption{Comparing FDE of methods using Bayesian signed-rank test. For PFSD and nuScenes, ROPE is defined as 0.5 meters. For SDD, ROPE is 1 pixel.}
\label{tab:baycomp2}
\scalebox{0.9}{
\begin{tabular}{c|c|c|c|c}
\toprule 
\multicolumn{5}{c}{\textbf{PFSD, nuScenes}}     \\
\midrule
% \multicolumn{5}{c}{Bayesian signed-rank test}     \\
% \midrule
Method A & p(A $>$ B)  & p(A $\approx$ B)  & p(A $<$ B)  & Method B \\
\midrule
T++     & 0.00     & 0.27    & \textbf{0.73}     & Y-Net       \\
T++     & 0.00     & 0.27    & \textbf{0.73}     & AF       \\
T++     & 0.00     & 0.27    & \textbf{0.73}     & Ours       \\
Y-Net   & 0.00     & 0.43    & \textbf{0.57}     & AF       \\
Y-Net   & 0.00     & 0.27    & \textbf{0.73}     & Ours       \\
AF      & 0.00     & \textbf{1.00}    & 0.00     & Ours       \\
% \midrule
% \multicolumn{5}{c}{Score}     \\
% \midrule
% Method & \# win & \# tie & \# lose & Rank \\
% \midrule
% T++    & 0      & 0      & 3       & 4    \\
% Y-net  & 1      & 0      & 2       & 3    \\
% AF     & 2      & 1      & 0       & 1.5  \\
% Ours   & 2      & 1      & 0       & 1.5 \\
\bottomrule
\end{tabular}
}

\vspace{1em}
\scalebox{0.9}{
\begin{tabular}{c|c|c|c|c}
\toprule 
\multicolumn{5}{c}{\textbf{SDD}}     \\
\midrule
% \multicolumn{5}{c}{Bayesian signed-rank test}     \\
% \midrule
Method A & p(A $>$ B)  & p(A $\approx$ B)  & p(A $<$ B)  & Method B \\
\midrule
T++     & 0.00     & 0.04    & \textbf{0.96}     & Y-Net       \\
T++     & 0.00     & 0.04    & \textbf{0.96}     & AF       \\
T++     & 0.00     & 0.04    & \textbf{0.96}     & Ours       \\
Y-Net   & 0.00     & \textbf{0.56}    & 0.44     & AF       \\
Y-Net   & 0.00     & \textbf{1.00}    & 0.00     & Ours       \\
AF      & 0.00     & \textbf{1.00}    & 0.00     & Ours       \\
% \midrule
% \multicolumn{5}{c}{Score}     \\
% \midrule
% Method & \# win & \# tie & \# lose & Rank \\
% \midrule
% T++    & 0      & 0      & 3       & 4    \\
% Y-net  & 1      & 2      & 0       & 2    \\
% AF     & 1      & 2      & 0       & 2  \\
% Ours   & 1      & 2      & 0       & 2 \\
\bottomrule
\end{tabular}
}
\end{table}

\begin{table}[tbp]
\caption{Comparing KDE NLL of methods using Bayesian signed-rank test. ROPE is 0 in this case.}
\label{tab:baycomp3}
\scalebox{0.9}{
\begin{tabular}{c|c|c|c|c}
\toprule 
% \multicolumn{5}{c}{Bayesian signed-rank test}     \\
% \midrule
Method A & p(A $>$ B)  & p(A $\approx$ B)  & p(A $<$ B)  & Method B \\
\midrule
T++     & \textbf{1.00}     & 0.00    & 0.00     & Y-Net       \\
T++     & \textbf{0.99}     & 0.00    & 0.01     & AF       \\
T++     & 0.00     & 0.00    & \textbf{1.00}     & Ours       \\
Y-Net   & 0.31     & 0.00    & \textbf{0.69}     & AF       \\
Y-Net   & 0.00     & 0.00    & \textbf{1.00}     & Ours       \\
AF      & 0.00     & 0.00    & \textbf{1.00}     & Ours       \\
% \midrule
% \multicolumn{5}{c}{Score}     \\
% \midrule
% Method & \# win & \# tie & \# lose & Rank \\
% \midrule
% T++    & 2      & 0      & 1       & 2    \\
% Y-net  & 0      & 0      & 3       & 4    \\
% AF     & 1      & 0      & 2       & 3   \\
% Ours   & 3      & 0      & 0       & 1  \\
\bottomrule
\end{tabular}
}
\end{table}

\begin{table}[tbp]
\caption{Comparing ECFL of methods using Bayesian signed-rank test. ROPE is 1\%.}
\label{tab:baycomp4}
\scalebox{0.9}{
\begin{tabular}{c|c|c|c|c}
\toprule 
% \multicolumn{5}{c}{Bayesian signed-rank test}     \\
% \midrule
Method A & p(A $>$ B)  & p(A $\approx$ B)  & p(A $<$ B)  & Method B \\
\midrule
T++     & 0.00     & 0.00    & \textbf{1.00}     & Y-Net       \\
T++     & 0.00     & 0.00    & \textbf{1.00}     & AF       \\
T++     & 0.00     & 0.00    & \textbf{1.00}     & Ours       \\
Y-Net   & 0.02     & 0.27    & \textbf{0.70}     & AF       \\
Y-Net   & 0.00     & 0.04    & \textbf{0.96}     & Ours       \\
AF      & 0.00     & 0.01    & \textbf{0.99}     & Ours       \\
% \midrule
% \multicolumn{5}{c}{Score}     \\
% \midrule
% Method & \# win & \# tie & \# lose & Rank \\
% \midrule
% T++    & 0      & 0      & 3       & 4    \\
% Y-net  & 1      & 0      & 2       & 3    \\
% AF     & 2      & 0      & 1       & 2   \\
% Ours   & 3      & 0      & 0       & 1  \\
\bottomrule
\end{tabular}
}
% \vspace{-0.2in}
\end{table}

\begin{table}[t]
\caption{Average rank of the four contrasted approached, based on the Bayesian Signed Rank pairwise test results in \cref{tab:baycomp1,tab:baycomp2,tab:baycomp3,tab:baycomp4}, across all measures. %\comsy{is this based on all measures (including average and stdev)? I feel that the average ranking is inconsistent with the numbers I calculated. See the xlsx in misc folder in overleaf for my data.}
% \comvp{maybe only show this table, not the ones with 'scores', wins, ties, losses in each test?}
}
\label{tab:avgrank}
\centering
\scalebox{0.9}{
\begin{tabular}{cc}
\toprule
Method & Average Rank of Bayesian Test Results \\
\midrule
T++    & 3.50     \\
Y-net  & 3.00  \\
AF     & 2.16 \\
Ours   & \textbf{1.33} \\
\bottomrule
\end{tabular}
}
\end{table}

% \comvp{what would be nice, if possible for supplement, is to have instances of very similar env layouts, but modify them such that they suggest largely unimodal goal predictions.  Eg, in fig(a), the right-hand turn is blocked by a wall.  And see if the models are able to understand that and make unimodal goal predictions.}

\section{Additional Qualitative Evaluations}
\label{sec:vis}

\begin{figure*}[t]
\centering
\begin{subfigure}[b]{0.4\textwidth} 
{\includegraphics[width=\textwidth]{figures/legend.jpg}}
\end{subfigure}
\vspace{-2em}
\begin{flushright}
\begin{subfigure}[b]{0.29\textwidth} 
{\includegraphics[width=\textwidth]{figures/path_legend.jpg}}
\vspace{-2.5em}
\end{subfigure}
\end{flushright}

\begin{subfigure}[b]{0.592\textwidth} 
{\includegraphics[width=\textwidth]{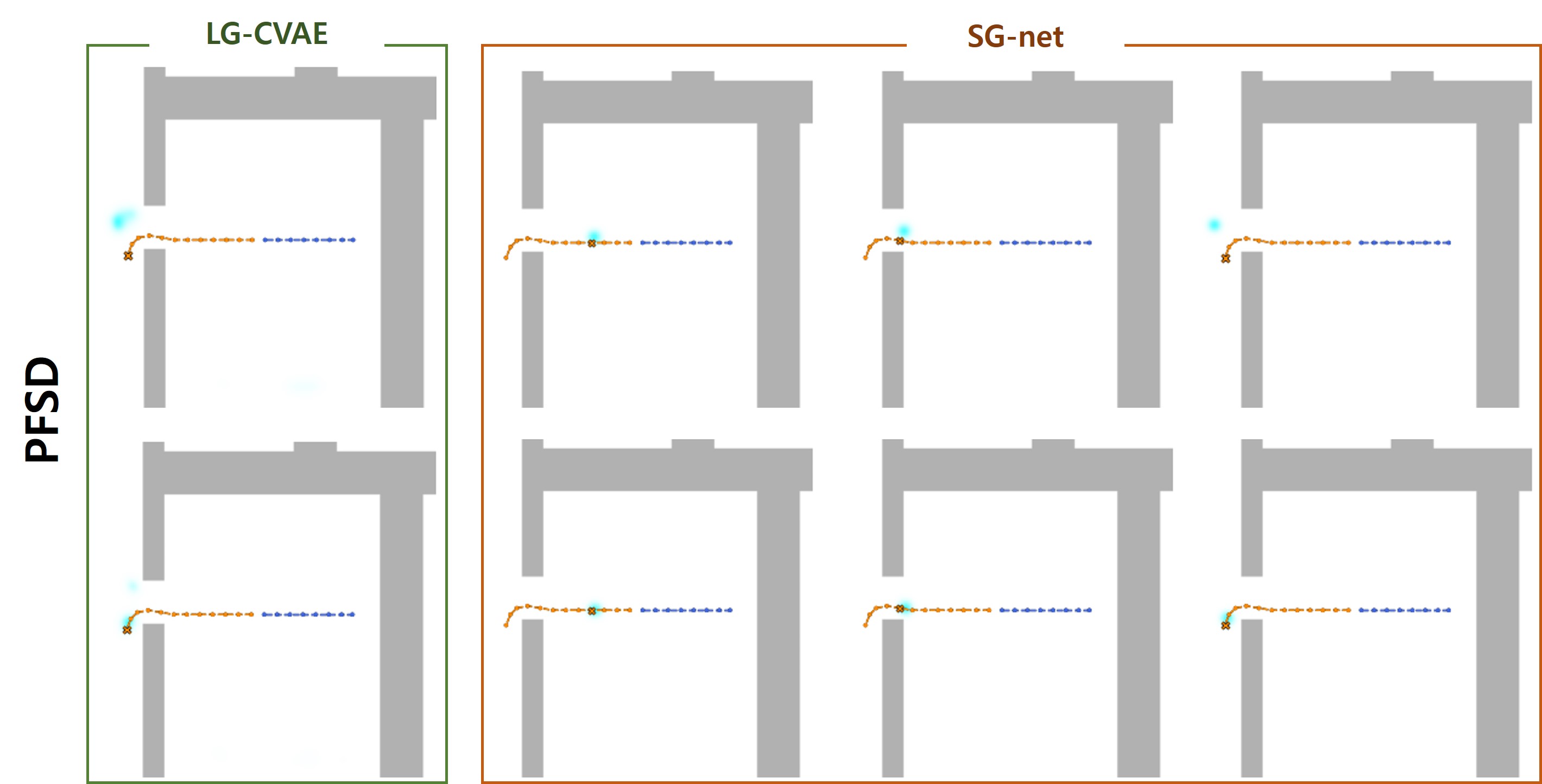} \caption{}\label{fig:macro_path2} }
\end{subfigure}
\hspace{3em}
\begin{subfigure}[b]{0.27\textwidth} 
{\includegraphics[width=\textwidth]{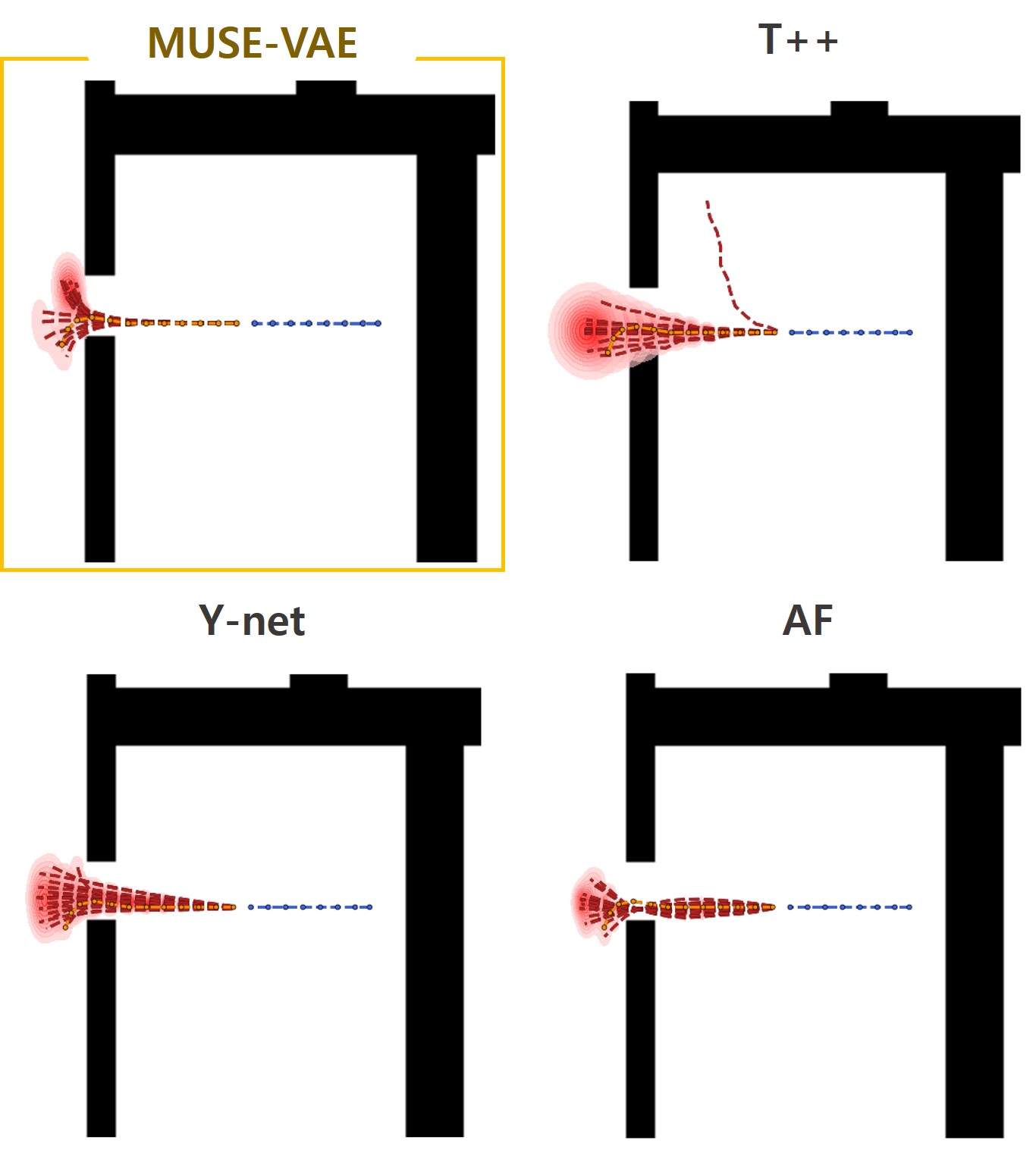} \caption{}\label{fig:micro_path2} }
\end{subfigure}

\vspace{-1.5em}
\begin{flushright}
\begin{subfigure}[b]{0.28\textwidth} 
{\includegraphics[width=\textwidth]{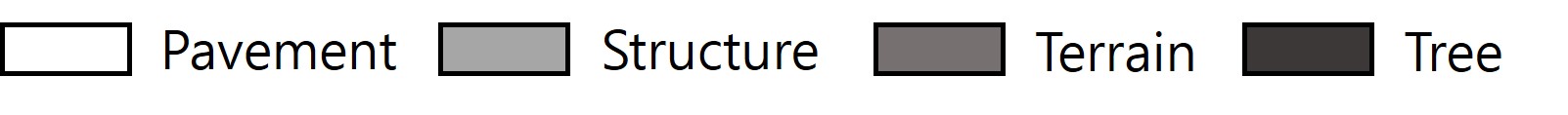}}
\vspace{-2.5em}
\end{subfigure}
\end{flushright}

\begin{subfigure}[b]{0.59\textwidth} 
{\includegraphics[width=\textwidth]{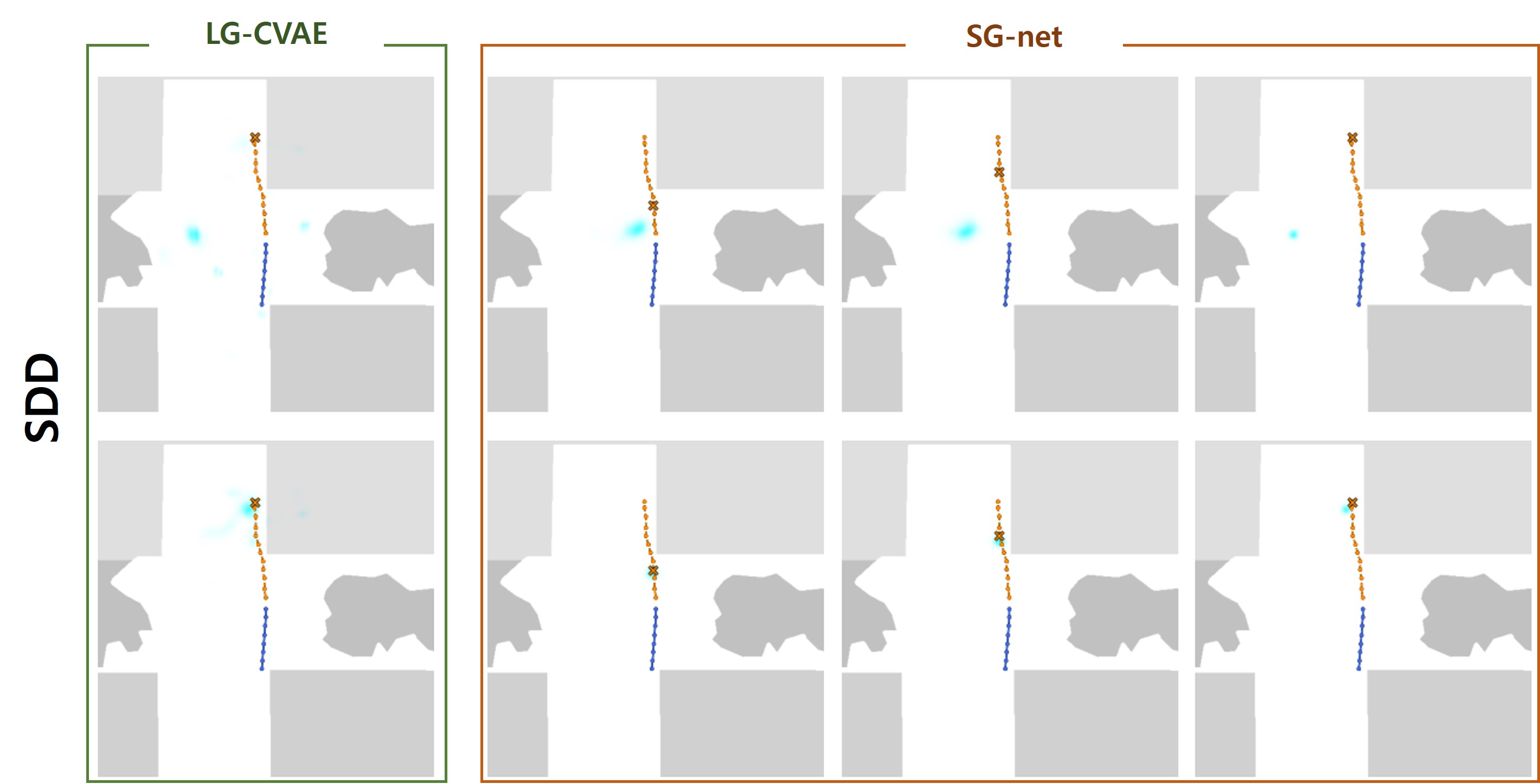} \caption{}\label{fig:macro_sdd2} }
\end{subfigure}
\hspace{3em}
\begin{subfigure}[b]{0.27\textwidth} 
{\includegraphics[width=\textwidth]{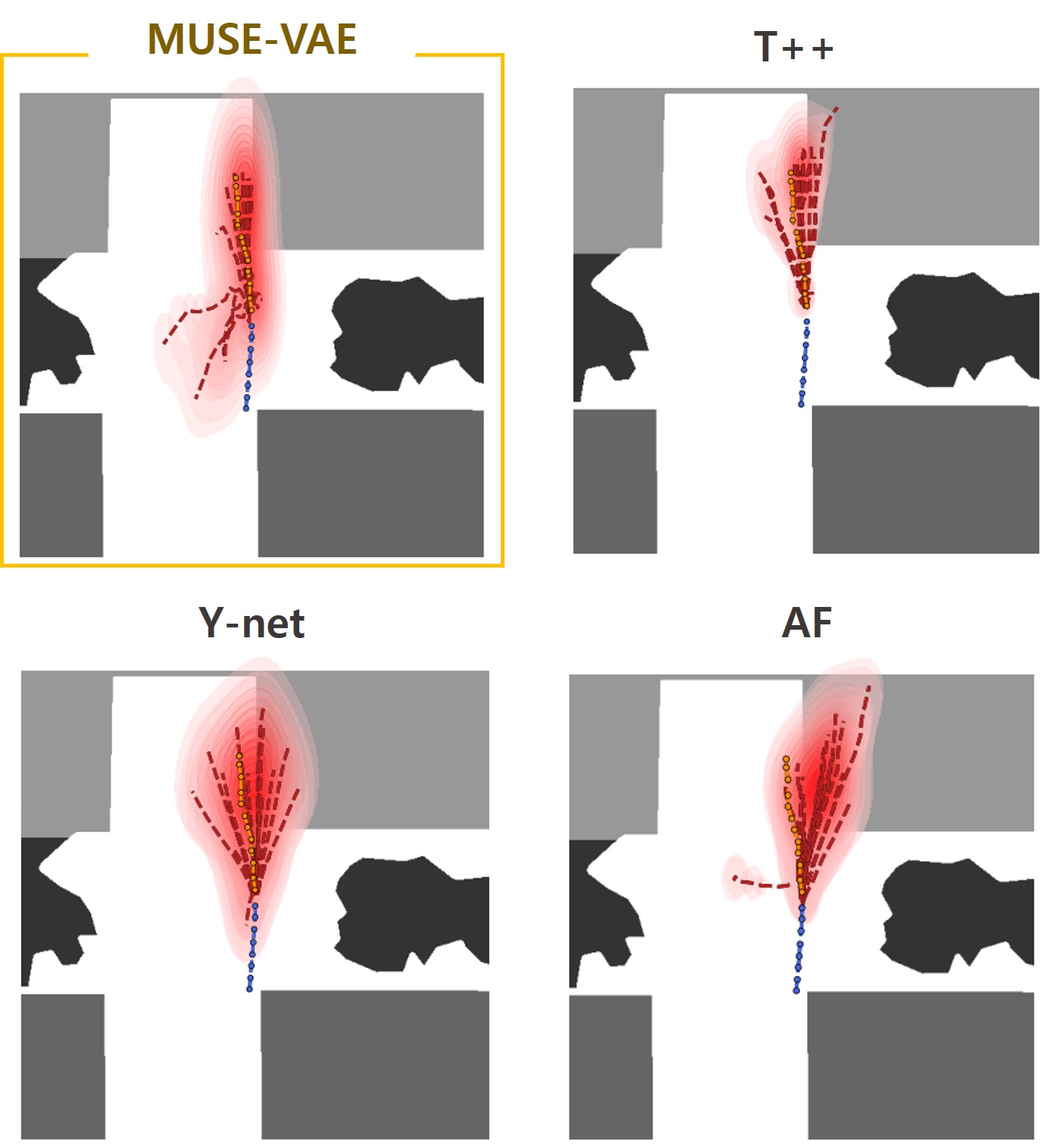} \caption{}\label{fig:micro_sdd2} }
\end{subfigure}

\vspace{-1.5em}
\begin{flushright}
\begin{subfigure}[b]{0.33\textwidth} 
{\includegraphics[width=\textwidth]{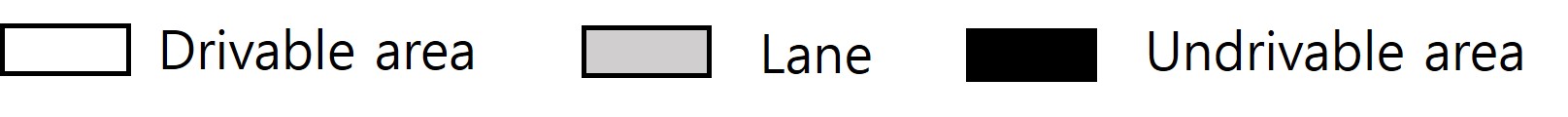}}
\vspace{-2.5em}
\end{subfigure}
\end{flushright}

\begin{subfigure}[b]{0.59\textwidth} 
{\includegraphics[width=\textwidth]{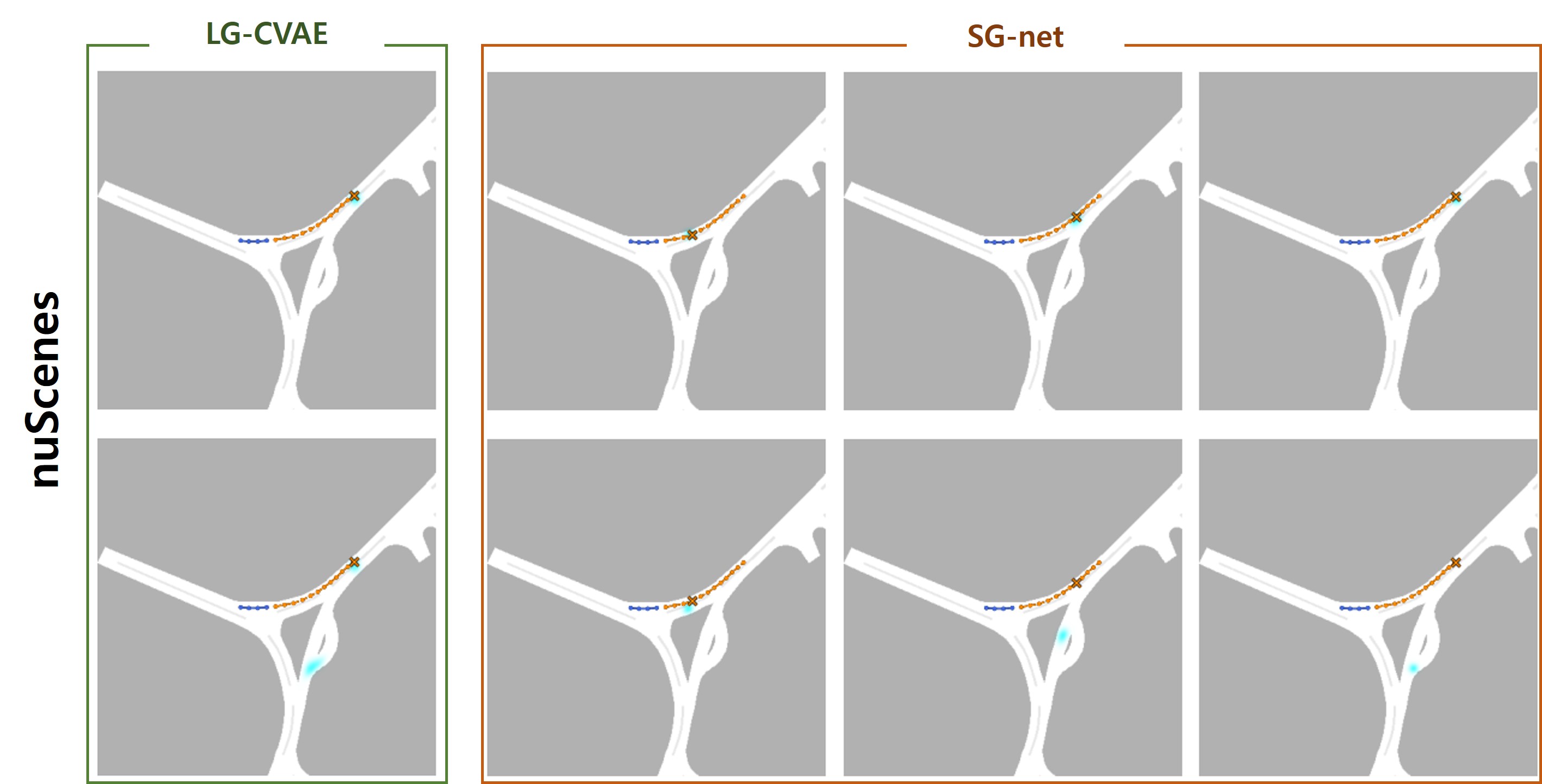} \caption{}\label{fig:macro_nu2} }
\end{subfigure}
\hspace{3em}
\begin{subfigure}[b]{0.27\textwidth} 
{\includegraphics[width=\textwidth]{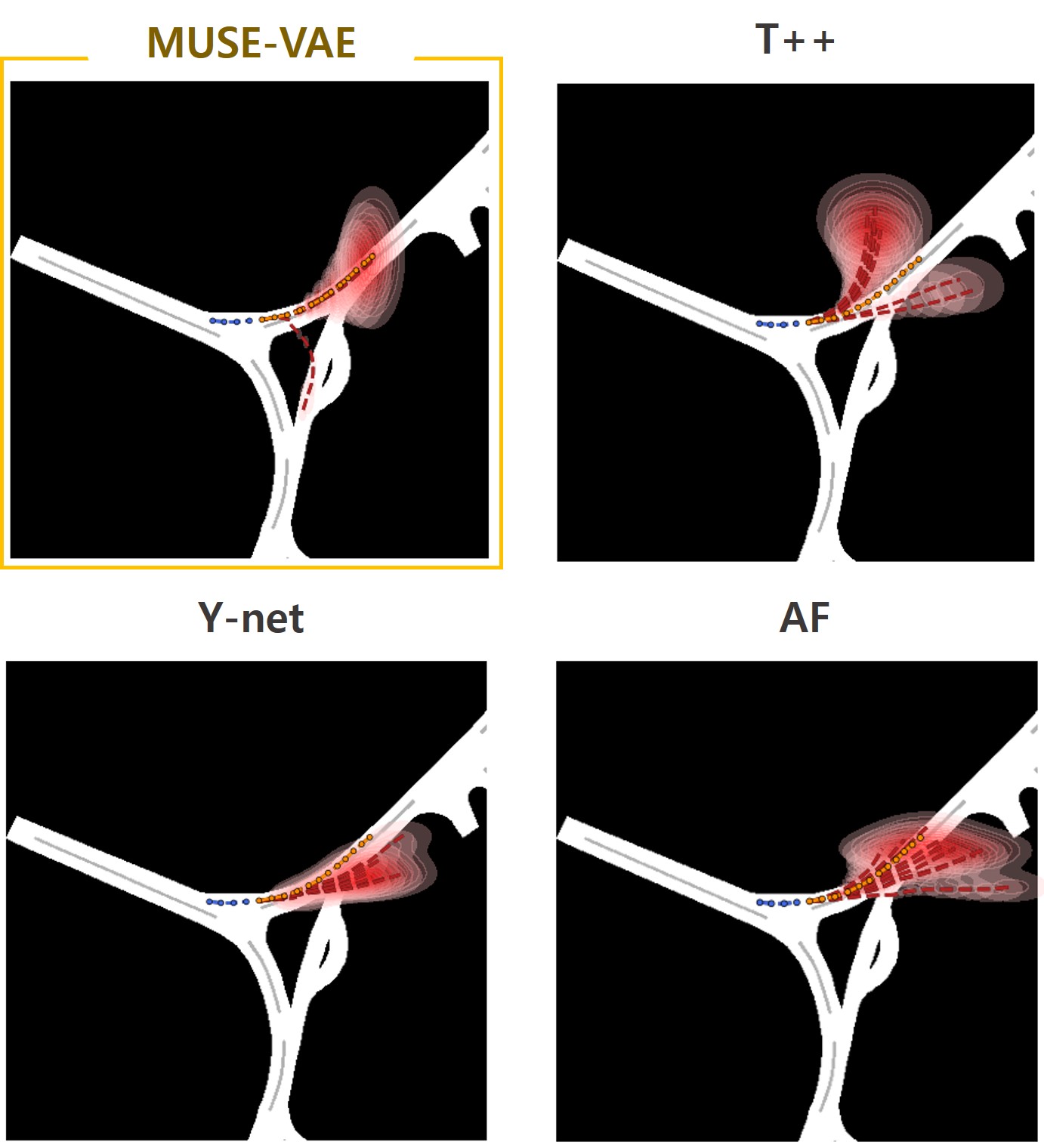} \caption{}\label{fig:micro_nu2} }
\end{subfigure}

\caption{
Left: Macro-stage results of (a) PFSD, (c) SDD, and (e) nuScenes respectively. In the first column, the Long-term Goal (LG) heat map prediction from LG-CVAE is overlaid on the local semantic map. The following three columns are two Short-term Goals (SG) and one LG from SG-Net. Here we show only two different sampling generations in each dataset.  The blue and orange lines indicate GT past and GT future trajectories, respectively. GT LG and SGs are marked with `x’. 
Right: Complete trajectory predictions of (b) PFSD, (d) SDD, and (f) nuScenes respectively. In each dataset, the 1st/2nd/3rd/4th image from top-left to bottom-right is  from Micro-stage of ours/Trajectron++/Y-net/AgentFormer, respectively. The blue, orange, and red lines indicate GT past, GT future, predicted future trajectories, respectively.
}
\label{fig:vis}
\end{figure*}

\autoref{fig:vis} shows qualitative results in the same manner as those presented in \autoref{fig:plot} of the Main paper. Here we investigate several key scenarios from each dataset, beyond the`fork-in-the-road' introduced in \autoref{fig:plot}. Scenarios were selected to highlight the challenges all models face in forecasting the environment-aware trajectories and offer insights into how the models behave when faced with environment constraints, in order to reveal the models' benefits and downsides. %qualitative evidence that the proposed \MUSE makes realistic trajectory predictions. 
% \comvp{could we report all quant metrics for each instance in the examples we show? (Best if we could show them in each plot.) If they agree with the qual assessment, they would strengthen our claims about Ours outperforming competitors}

The difference in the environment configurations between the two PFSD instances, \autoref{fig:micro_path2} here and \autoref{fig:micro_path} in the Main paper, is that \autoref{fig:micro_path2} has no obstacles in the direction of the observed, past trajectory while \autoref{fig:micro_path} presents obstacles at the bottom of the map in the same direction.
% \comvp{not sure what that means}
Thus, comparing the predictions, our method predicted both straight ahead and left or right curved trajectories for \autoref{fig:micro_path2}, while producing only left or right curves for \autoref{fig:micro_path}. %Similar trends are observable in other models. 
%\comvp{... suggesting what? If all models do well here, why is this showcasing our model's benefits?}
%\comsy{anything to say for fig.e1a?}

% \autoref{fig:micro_sdd2} shows an example when the past trajectories are mainly moving to the left side of the local map while slightly changing toward the top of the local map. The environment has no obstacles on the left side and there is another (navigable) pavement area at the top of the map. On the other hand, in \autoref{fig:micro_sdd}, the past trajectories show a more pronounced shift towards the lower left because the agent recognizes and tries to avoid the obstacle on the left side of the map. As a result, in \autoref{fig:micro_sdd2}, \MUSE predicts more going left than up, whereas \autoref{fig:micro_sdd} predictions of \MUSE tend to turning to the bottom side.
% As analyzed in the main paper, Y-net \cite{ynet} produces trajectories with various horizons that do not reflect the velocity of the past trajectories because of their test time sampling trick. In addition, the characteristic of the past trajectory of walking in a zigzag form is not observable in Y-net and AgentFormer \cite{AgentFormer} since their predictions are made based on their diversity strategy rather than from the learned distribution. On the other hand, T++ and \MUSE can reflect this and provide not only straight predictions but also gait that naturally moves back and forth.
%\comsy{anything to say for fig.e1c?}
\autoref{fig:micro_sdd2} shows an example when the ground truth trajectory passes right next to the `structure' area, which is non-navigable; the heading direction is mostly blocked by the structure. Our model can make predictions that do not violate the environmental constraints, going back or turning left to search for navigable space. On the other hand, the predictions from the baseline models collide with the obstacles, a violation of the desired behavior.

\autoref{fig:micro_nu2} is a fork-in-the-road scenario like \autoref{fig:micro_nu}, with another drivable area on the other side of the fork in the road. 
Although the traffic flow in this area is in the direction opposite to the predicted trajectory, it still is a drivable area.
%Although this area is a road for the \comvp{reverse progress of the current agent's direction, it is true that it is also drivable - I don't understand this sentence}. 
Since we have never provided a clear guidance for learning in which direction to drive based on the `correct' lane, our model simply treats this area as drivable and makes one possible prediction.
It can be seen that the baseline models cannot consider this possibility, instead making many predictions into the undrivable area.

%\comsy{I think this would be a nice place to report the semantic violation metric since if we still beat other methods with that metric, we can argue that our method's trajectory is realistic albeit this seemingly unrealistic case}

Finally, \autoref{tab:vis} shows the corresponding quantitative results for each dataset with metrics introduced in \autoref{sec:metric}. The results in the table are well-aligned with the visualization in the \autoref{tab:vis}. \MUSE shows the highest ECFL in all datasets, suggesting our model forecasts environmentally-compliant trajectories. Moreover, our model shows the best performance for all datasets in terms of ADE; it similarly leads in FDE performance in SDD and nuScenes.  \MUSE attains the second best results in FDE for PFSD, trailing the top Y-Net by only 0.01 meter. Similarly, \MUSE approaches the top method (AF) in KDE NLL for nuScenes. The third ranked performance of our model in KDE NLL of SDD stems from those predictions heading to the left or going back toward the past trajectories. While away from the specific trajectory taken by the agent in this instance, the behaviors predicted by \MUSE are very reasonable strategies for an agent who reaches a dead-end.

% We solidify the validity of each model's degree of the environment-compliance with another evaluation metric suggested in \cite{Zhu2021MotionFW}. 
% \todo{Sam, please explain the difference between ECFL and the new metric.}

\begin{table}[t]
\caption{Quantitative results of \autoref{fig:vis}. 
 PFSD and SDD with $t_p=3.2$s (8 frames) and $t_f=4.8$s (12 frames), and nuScenes with $t_p=2$s (4 frames) and $t_f=6$s (12 frames). Errors are in meters for PFSD and nuScenes, and in pixels for SDD.
% \comsy{anything to comment?}
}
\label{tab:vis}
\vspace{-1.5em}
\begin{center}
\scalebox{0.85}{
\begin{tabular}{c|c|c|c|c|c}
\toprule
Dataset& Model & ADE $\downarrow$ & FDE $\downarrow$ & KDE NLL $\downarrow$ & ECFL $\uparrow$  \\
\midrule
\midrule

& T++
& 0.16 & 0.05 & -1.54 & 95 \\
\multirow{2}{*}{PFSD}
                    & Y-net                   
& 0.1 & \textbf{0.04} & -0.76 & \textbf{100} \\
\multirow{2}{*}{($K$ = 20)} 
                    & AF  
& 0.12 & 0.05 & -0.50 & \textbf{100} \\
                    & Ours 
& \textbf{0.08 }& 0.05 &\textbf{ -4.24 }& \textbf{100} \\
\midrule

& T++            
& 4.15 & 3.18 & \textbf{6.58} & 80 \\
\multirow{2}{*}{SDD} 
                    & Y-net                   
& 3.15 & 2.88 & 7.77 & 65 \\
\multirow{2}{*}{($K$ = 20)} 
                    & AF  
& 13.44 & 7.10 & 8.69 & 20 \\

                    & Ours                    
& \textbf{2.86} & \textbf{2.34} & 8.45 & \textbf{100} \\       
\midrule

& T++            
& 4.92 & 4.91 & 3.91 & 0\\
\multirow{2}{*}{nuScenes} 
                    & Y-net                   
& 3.08 & 2.99 & 6.21 & 40 \\
\multirow{2}{*}{($K$ = 10)} 
                    & AF  
& 1.17 & 1.08 & \textbf{3.68} & 30 \\
                    & Ours                    
& \textbf{0.89} & \textbf{0.73} & 3.84 & \textbf{90 }\\    
\bottomrule

\end{tabular}}
\end{center}
\vspace{-2em}
\end{table}

% \begin{table}[t]
% \caption{Context-Violation-Rate on nuScenes
% }
% \label{tab:viol}
% \centering
% \scalebox{0.9}{
% \begin{tabular}{c|c|c}
% \toprule

% K&Method & Context-Violation-Rate \\
% \midrule
% \midrule
% \multirow{4}{*}{5}
% & T++
% & \\
%                     & Y-net                   
% & \\
%                     & AF  
% & \\
%                     & Ours 
% & \\
% \midrule
% \multirow{4}{*}{10}
% & T++
% & \\

%                     & Y-net                   
% & \\
%                     & AF  
% & \\
%                     & Ours 
% & \\

% \bottomrule
% \end{tabular}
% }
% \end{table}

\section{Limitation of SDD Segmentation}
\label{sec:sdd_seg}
\begin{figure}
\centering
\begin{subfigure}[tbhp]{0.45\textwidth}
 {\includegraphics[width=\textwidth]{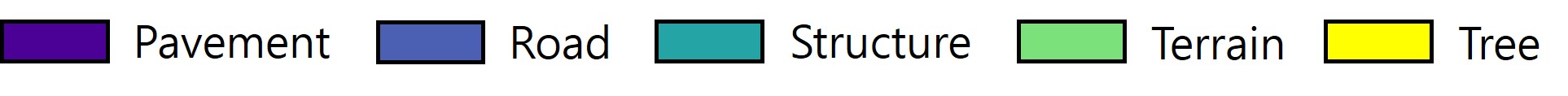} 
 }
 \vspace{-1em}
\end{subfigure}

\begin{subfigure}[tbhp]{0.45\textwidth}
 {\includegraphics[width=\textwidth]{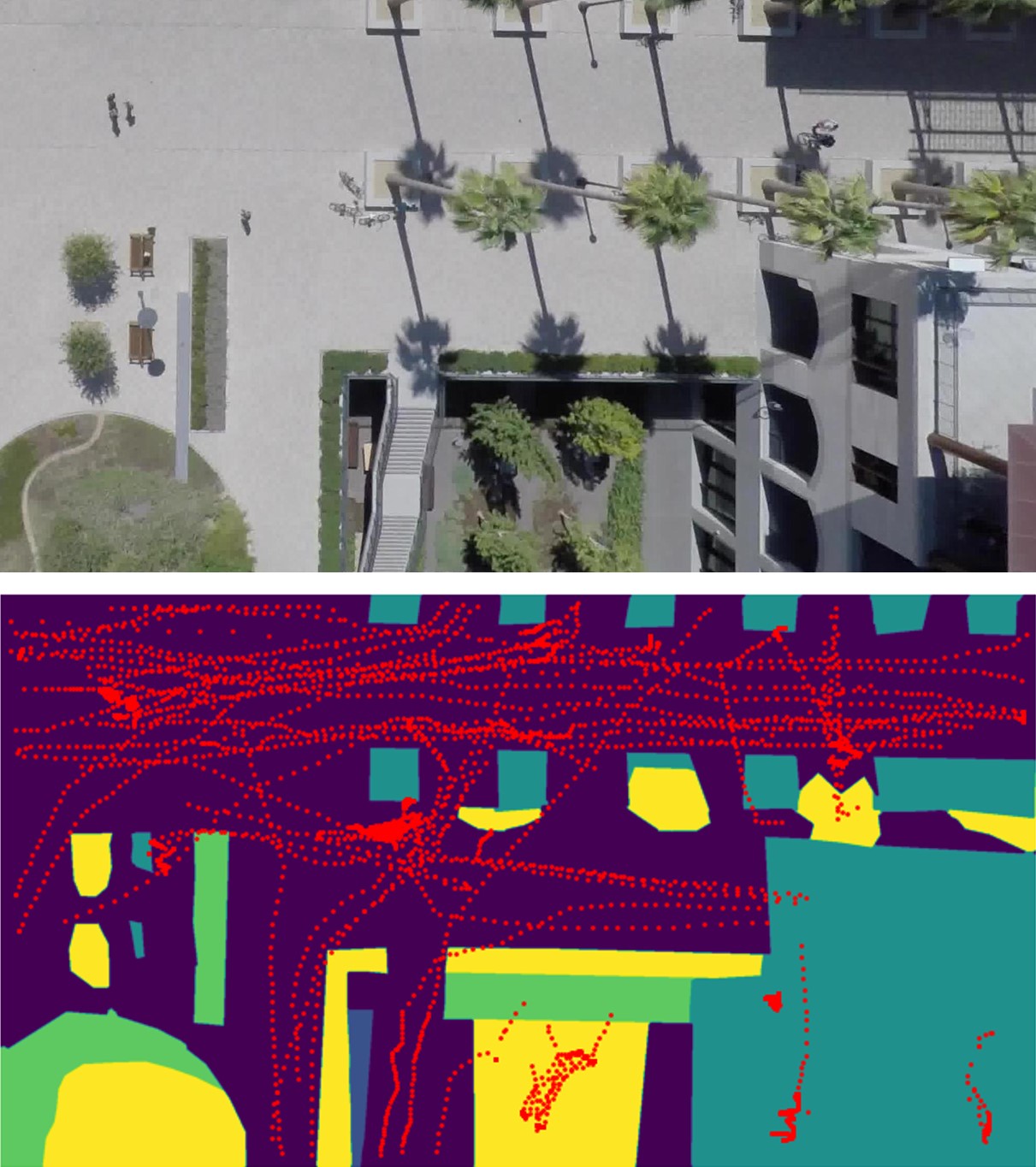} 
 \caption{coupa\_0} \label{fig:coupa} }
\end{subfigure}

\begin{subfigure}[tbhp]{0.45\textwidth}
 {\includegraphics[width=\textwidth]{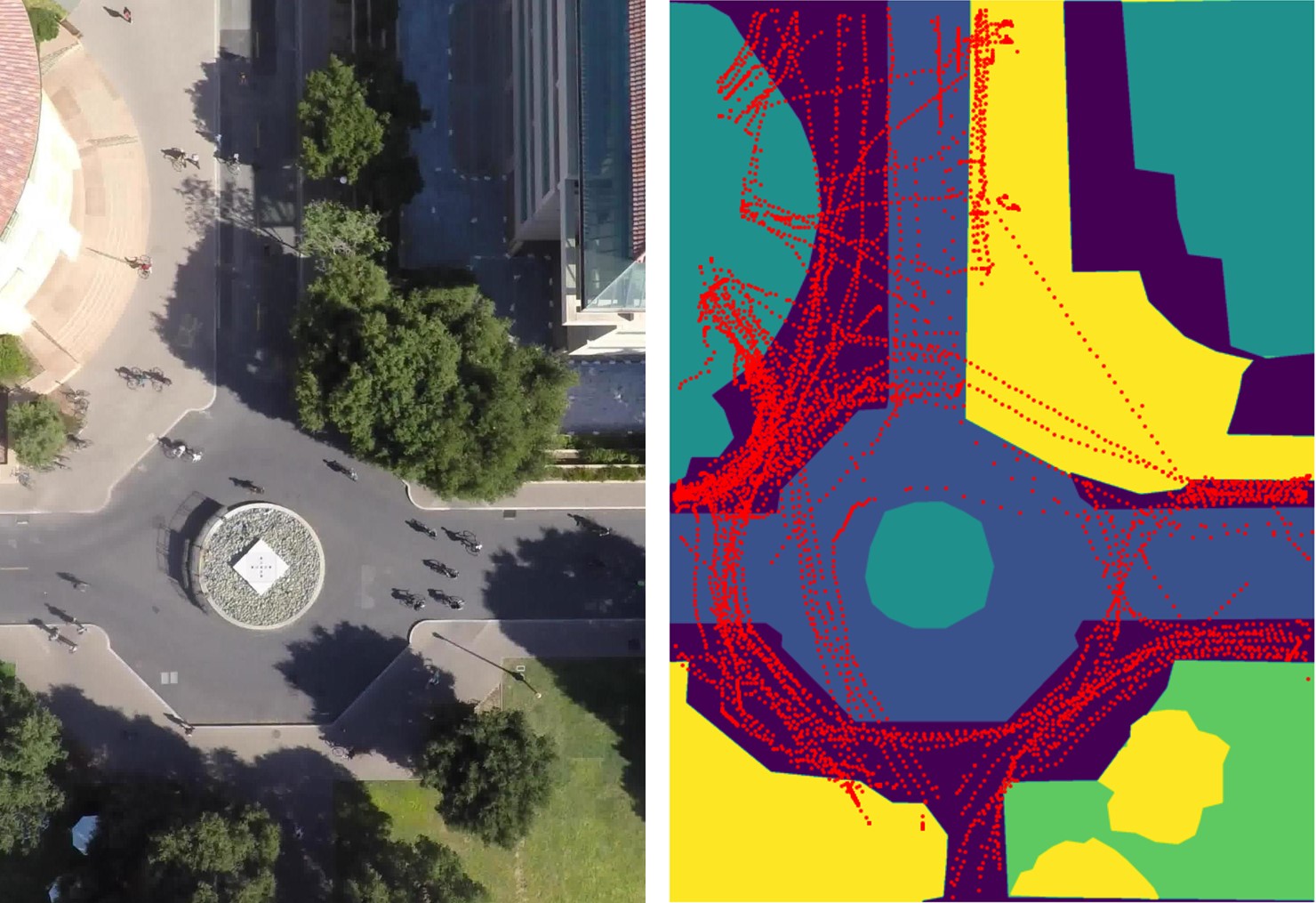} 
 \caption{little\_3} \label{fig:little} 
 }
\end{subfigure}
\vspace{-0.1in}
  \caption{ 
  SDD scene image and its semantic map of the scene (a) coupa\_0 and (b) little\_3. Red points indicate all trajectories in each scene. Trajectories are found in the region with classes like `structure' or `tree', which is unexpected in terms of navigability.
%   \comsy{anything to comment? you can reiterate or emphasize what's described in the text}
}
\label{fig:sdd_seg}
\vspace{-1.5em}
\end{figure}

In our evaluations, we used the semantic map of SDD provided by Y-net. They classify the scene environment into the five classes described in \autoref{sec:impl}. In \autoref{fig:sdd_seg}, we show SDD scene images and their semantic maps of the scene (a) coupa\_0 and (b) little\_3. Red points indicate all trajectories in each scene. 

There are two major problems in learning this map. First, it is not clear which semantic classes ought to be considered as navigable. Based on the class names, only pavement and road may be reasonably navigable space, but as seen in \autoref{fig:sdd_seg}, there are trajectories on tree, terrain, and structure. For the evaluation, we set only the `structure' class as the obstacle class. Secondly, the segmentation regions are semantically inaccurate. Near the bottom-center of the semantic map of \autoref{fig:coupa}, we can see the squares all colored in yellow, which indicates a tree. However, looking at the scene image, we notice that not all of those regions are trees. 

These inaccurate annotations give rise to the model confusion on how to deal with the map information when determining the trajectories that should only exist in navigable spaces. This affects \MUSE more significantly than other models since the decision of long-term and short-term goals in Macro-stage heavily depend on the local map information; this subsequently leads to slightly lower ECFL for SDD in \autoref{sec:quan} of the Main paper, compared to other approaches.
% \comsy{so what? reiterate the point mentioned in the main paper why this is something reviewers need to know}

\section{Challenges and Future Work}
\label{sec:challenges}
In this paper, we proposed to ``boost" the learning of models that forecast realistic, environment-aware trajectories by leveraging the large body of scene-compliant simulated trajectories in PFSD, a complex environment with intricate navigable / non-navigable structures. These structures were designed to induce diverse agent-environment behaviors, hence data to train models, that generalize well to many real-world scenarios.

Another component that makes trajectory prediction realistic is the absence of collisions among the agents themselves. One way to learn models that accomplish this, aside from collecting large bodies of real-world data, is to create synthetic datasets that reflect the desired agent-agent relationships, much like PFSD captures the environment-agent interactions. Such trained models would then transfer to the (smaller) real world datasets.

However, synthesizing collision-free models for agent-agent interactions is a challenging task. While designing scenarios where only the inter-agent distance is kept above a certain threshold is possible, such instances directly eliminate more complex yet desired behaviors such as agents walking together as a group or agents passing by each other in the opposite directions.  
%. However, if a model simply maintains a certain distance from other agents, it can bring a trade-off that reduces the accuracy of individual trajectory prediction itself because the amount of agent-agent distance has social meanings. Therefore, it is important to learn agent-relation by understanding the context with surrounding agents. A model needs to take into account whether the agents are walking together as a group or are merely passing by each other at the moment in the forward / reverse direction.
Moreover, the behavioral patterns determining inter-agent proximity are also contextualized by the surrounding environment. For instance, the density of agents (hence their mutual displacements), will be higher (displacements lower) in very narrow navigable spaces compared to those in wider, open environments.

We leave it as an open research challenge to study such integrated models that can consider the inter-agent relationship in addition to the agent-environment interactions we tackled here using our \MUSE.

%%%%%%%%% REFERENCES
\clearpage

{\small
\bibliographystyle{ieee_fullname}
\bibliography{main}
}

\end{document}